\documentclass{article}
\PassOptionsToPackage{numbers}{natbib} 

\usepackage[main, final]{neurips_2026}

\usepackage[utf8]{inputenc} 
\usepackage[T1]{fontenc}    
\usepackage{hyperref}       
\usepackage{url}            
\usepackage{booktabs}       
\usepackage{amsfonts}       
\usepackage{nicefrac}       
\usepackage{microtype}      
\usepackage{xcolor}         

\usepackage{amsmath}
\usepackage{amssymb}
\usepackage{mathtools}
\usepackage{amsthm}
\usepackage{dsfont}
\usepackage{lipsum}
\usepackage{bm,bbm}
\usepackage{bbding}
\usepackage{paralist}
\usepackage{xspace}
\usepackage[utf8]{inputenc} 
\usepackage[T1]{fontenc}    
\usepackage{hyperref}       
\usepackage{url}            
\usepackage{amsfonts}       
\usepackage{nicefrac}       
\usepackage{microtype} 
\usepackage{multirow}
\usepackage{graphicx}
\usepackage{ragged2e}
\usepackage[most]{tcolorbox}
\definecolor{darkerlogocolor}{RGB}{20, 0, 145}
\newtcolorbox{ttcolorbox}[1][]{colframe=darkerlogocolor, colback=darkerlogocolor!4!white, title=#1}
\usepackage[font=small,labelfont=bf]{caption}
\usepackage{subfloat}
\usepackage{float}
\usepackage{algorithm,algorithmic}
\usepackage{wrapfig}
\usepackage{booktabs}
\usepackage{verbatim}
\usepackage{setspace}
\usepackage{tikz}
\usepackage{tcolorbox}
\definecolor{mydarkblue}{rgb}{0,0.08,0.45}
\hypersetup{
    colorlinks,
    breaklinks,
    linkcolor = mydarkblue,
    citecolor = mydarkblue,
    urlcolor  = black,
} 
\def \E {\mathbb{E}}
\def \x {\mathbf{x}}

\def \D {\mathcal{D}}

\def \O {\mathcal{O}}

\def \q {\mathbf{q}}

\def \q {\mathbf{q}}

\def \B {\mathbf{B}}

\def \B {\mathcal{B}}

\usepackage{mathtools}
\usepackage{appendix}

\newtheorem{myThm}{Theorem}

\newtheorem{myLemma}{Lemma}

\newtheorem{myProp}{Proposition}

\theoremstyle{definition}
\newtheorem{myAssum}{Assumption}

\newtheorem{myCor}{Corollary}

\newtheorem{myRemark}{Remark}

\def \E {\mathbb{E}}
\def \D {\mathcal{D}}

\def \x {\mathbf{x}}

\def \epsilon {\varepsilon}

\definecolor{DSgray}{cmyk}{0,1,0,0}

\usepackage[textsize=tiny]{todonotes}
\usepackage{prettyref}

\newcommand{\savehyperref}[2]{\texorpdfstring{\hyperref[#1]{#2}}{#2}}
\newrefformat{eq}{\savehyperref{#1}{Eq.~(\ref*{#1})}}
\newrefformat{eqn}{\savehyperref{#1}{Equation~\ref*{#1}}}
\newrefformat{lem}{\savehyperref{#1}{Lemma~\ref*{#1}}}
\newrefformat{lemma}{\savehyperref{#1}{Lemma~\ref*{#1}}}
\newrefformat{def}{\savehyperref{#1}{Definition~\ref*{#1}}}
\newrefformat{line}{\savehyperref{#1}{Line~\ref*{#1}}}
\newrefformat{thm}{\savehyperref{#1}{Theorem~\ref*{#1}}}
\newrefformat{corr}{\savehyperref{#1}{Corollary~\ref*{#1}}}
\newrefformat{cor}{\savehyperref{#1}{Corollary~\ref*{#1}}}
\newrefformat{sec}{\savehyperref{#1}{Section~\ref*{#1}}}
\newrefformat{app}{\savehyperref{#1}{Appendix~\ref*{#1}}}
\newrefformat{appendix}{\savehyperref{#1}{Appendix~\ref*{#1}}}
\newrefformat{assum}{\savehyperref{#1}{Assumption~\ref*{#1}}}
\newrefformat{ex}{\savehyperref{#1}{Example~\ref*{#1}}}
\newrefformat{fig}{\savehyperref{#1}{Figure~\ref*{#1}}}
\newrefformat{alg}{\savehyperref{#1}{Algorithm~\ref*{#1}}}
\newrefformat{rem}{\savehyperref{#1}{Remark~\ref*{#1}}}
\newrefformat{conj}{\savehyperref{#1}{Conjecture~\ref*{#1}}}
\newrefformat{prop}{\savehyperref{#1}{Proposition~\ref*{#1}}}
\newrefformat{proto}{\savehyperref{#1}{Protocol~\ref*{#1}}}
\newrefformat{prob}{\savehyperref{#1}{Problem~\ref*{#1}}}
\newrefformat{claim}{\savehyperref{#1}{Claim~\ref*{#1}}}
\newrefformat{que}{\savehyperref{#1}{Question~\ref*{#1}}}
\newrefformat{op}{\savehyperref{#1}{Open Problem~\ref*{#1}}}
\newrefformat{fn}{\savehyperref{#1}{Footnote~\ref*{#1}}}
\allowdisplaybreaks

\usepackage{subcaption}
\def \rmsrc {\mathrm{src}}
\def \rmtgt {\mathrm{tgt}}
\hypersetup{hidelinks}

\title{Dataset Mismatch Matters in Group Relative Policy Optimization for Reinforcement Learning\\ from Verifiable Rewards}

\author{%
  Zhen-Yu Zhang \\
  RIKEN-AIP\\
  \texttt{zhen-yu.zhang@riken.jp} \\
  \And
  Jiandong Zhang \\
  Northeastern University\\
  \texttt{zhang.jiando@northeastern.edu} \\
  \And
  Huaxiu Yao \\
  UNC-Chapel Hill\\
  \texttt{huaxiu@cs.unc.edu} \\
  \And
  Gang Niu \\
  RIKEN-AIP\\
  \texttt{gang.niu.ml@gmail.com} \\
  \And
  Masashi Sugiyama \\
  RIKEN-AIP\\
  The University of Tokyo\\
  \texttt{sugi@k.u-tokyo.ac.jp} \\
}

\begin{document}

\maketitle

\begin{abstract}
Recently, \emph{reinforcement learning from verifiable rewards}~(RLVR) is practically very important, where the leading approach is \emph{group relative policy optimization}~(GRPO) that typically faces a trade-off between rollout efficiency and downstream performance.
To improve efficiency while preserving performance, existing \emph{selective rollout} methods mainly exploit \emph{only source-side} training dynamics to optimize the discrete distribution over source prompts.
However, in practice, the source training dataset is rarely identically distributed to the target test dataset.
When such a mismatch exists, due to legal or privacy constraints, it is usually not acceptable to share target prompts themselves with the RLVR trainer.
Nevertheless, it is still possible for the target data owner to provide limited aggregated feedback from a small held-out target set without revealing target prompts or rollout traces.
In this paper, we propose \emph{learning-to-reweight rollout}~{(L2RR)} that explicitly incorporates target-side performance feedback into GRPO.
Specifically, L2RR learns a discrete sampling distribution over source prompts by optimizing a bi-level objective that maximizes accuracy on a held-out target dataset, while simultaneously penalizing rollouts that provide little incremental learning information.
Empirical studies on six math reasoning benchmarks and three model scales show that L2RR improves target-domain performance and increases rollout efficiency.
\end{abstract}


\section{Introduction}
\label{sec:intro}

Recent \emph{large language models}~(LLMs), such as OpenAI o1~\citep{jaech2024openai} and DeepSeek R1~\citep{guo2025deepseek}, have achieved strong reasoning performance across a variety of domains~\citep{hendrycks2measuring,rein2024gpqa}, since they leverage increased inference-time computation for reasoning.
A major catalyst behind this progress is \emph{reinforcement learning with verifiable rewards}~(RLVR), which optimizes policies from binary-valued verifiable feedback~\citep{lambert2024tulu,guo2025deepseek}.
\emph{Group relative policy optimization}~(GRPO) is a representative RLVR algorithm that trains via verifier-based rollouts, where a prompt is sampled, a group of responses is generated and scored, and the policy is updated using \emph{group-normalized advantages}~\citep{shao2024deepseekmath}.
When group rewards collapse to the same value, the resulting advantages become near zero, so obtaining non-zero advantages often requires many costly rollouts.
This creates a fundamental trade-off: fewer rollouts reduce cost but can degrade downstream performance, whereas better performance often requires more rollouts.
Therefore, \emph{improving rollout efficiency while preserving downstream performance} is critical for scaling RLVR training.

Existing methods improve rollout efficiency using training-side GRPO dynamics~\citep{yu2025dapo,zheng2025act}.
Recent \emph{selective rollout} methods leverage temporal regularities to either skip such prompts before expensive rollouts~\citep{zheng2025act} or reuse previously successful rollouts to reduce repeated sampling~\citep{zhang2025improving}.
These rollout-selection strategies rely on training-side information.
Consequently, their efficiency gains may be limited by what the training dataset can reveal, and rollout-or-not decisions are tuned using training-side validation, in the hope that these choices transfer to downstream performance.

However, in practice, \emph{the source training dataset is rarely identically distributed to the target test dataset}.
Even in math reasoning, the training and test datasets can differ substantially in their difficulty or subdomains, with test problems often skewing toward harder or covering different topics~\citep{wang2024embedding}.
As a result, source-side GRPO dynamics can yield rollout decisions that are misaligned with target performance, motivating the use of target information to learn selective rollout strategies.

Note that leveraging target-side information is non-trivial.
In many real-world tasks, the model owner and the data owner are different entities, so target-side prompts and rollout traces can be inaccessible for RLVR training because of privacy or legal constraints~\citep{ye2024fedllm,bai2024federated}.
Without any target information, improvements beyond source-only training may be limited.
Fortunately, in many real-world scenarios, although the trainer may not be allowed to access target prompts or target-side rollout traces, it is often still possible to access \emph{limited target-side evaluation feedback} computed on a small held-out target dataset.
Depending on the deployment protocol, this feedback may take the form of aggregated target-side gradients or directional derivatives, without directly sharing target prompts or rollout traces.
For example, in hospitals, target prompts stay on-premise and training proceeds via fine-tuning with only model updates shared~\citep{bai2024federated}; similarly, some benchmark-style workflows provide aggregate scores through a remote evaluator, without releasing test inputs during blind evaluation~\citep{guyon2019analysis}.
This motivates our core question: \emph{can we leverage such limited target-side information to improve selective-rollout efficiency and target generalization?}

In this paper, we propose \emph{learning-to-reweight rollout}~{(L2RR)}, a target-guided selective rollout scheme for GRPO that uses target-side evaluation feedback to learn how rollouts are allocated across the source prompt pool.
Specifically, L2RR learns a discrete distribution for sampling over source prompts and alternates between GRPO policy updates and sampler updates.
The GRPO policy updates are as usual.
The sampler updates adjust prompt sampling probabilities using gradient-based feedback from the held-out target objective and the frequency of yielding near-zero group-normalized advantages.
Benefiting from this design, L2RR jointly prioritizes target-relevant rollouts and avoids repeatedly spending resources on prompts that contribute little to learning progress, using only limited target-side information.
We provide a theoretical analysis establishing an $\O(T^{-1/2})$ target-loss bound up to bias and tracking residuals under regularity assumptions and an appropriate choice of regularization weight, where $T$ is the training horizon.
Empirical studies on six reasoning benchmarks and three models spanning two parameter scales show that L2RR improves target performance while simultaneously increasing rollout efficiency.

Although \emph{learning-to-reweight}~\citep{ren2018learning}\footnote{\cite{shu2019meta} further learned a weighting function that generalizes to new data, whereas our setting does not require this.} shares a similar idea and has been studied in supervised learning settings, extending this paradigm to RLVR is non-trivial.
GRPO training dynamics are time-varying, and the advantage of a prompt can degenerate as group rewards collapse, so rollout effectiveness is required to be modeled using history-dependent information during training.


\section{Preliminary}
\label{subsec:preliminary}

In this section, we first introduce the notation and problem setup, and then briefly review the GRPO training procedure, along with recent approaches for mitigating advantage collapse in GRPO.

\textbf{Notations.} Let $\D_{\rmsrc} = \{(\x^{\mathrm{S}}_i, a^{\mathrm{S}}_i)\}_{i=1}^{N_S}$ be the source dataset and locally held validation set $\D_{\rmtgt} = \{(\x^{\mathrm{T}}_i,a^{\mathrm{T}}_i)\}_{i=1}^{N_T}$ the target dataset, where $\x^{\mathrm{S}}_i$ and $\x^{\mathrm{T}}_i$ are prompts and $a^{\mathrm{S}}_i$ are the corresponding answers of $\x^{\mathrm{S}}_i$.
We consider a training procedure of $T$ rounds.
For each prompt $\x$, at round $t\in\{1,\ldots,T\}$, the LLM policy $\pi_{\theta_t}(y \mid \x)$ generates an output $y$.
A verifier evaluates the output and returns a binary reward $r(\x, y, a)\in\{0,1\}$, indicating whether the final answer matches the reference answer $a$.

\textbf{Group relative policy optimization.} In GRPO~\citep{shao2024deepseekmath}, at each round $t\in\{1, \ldots, T\}$, we sample a batch of prompts-answer pairs $\B_t$ of size $B$.
For each sampled prompt $(\x^{\mathrm{S}},a^{\mathrm{S}}) \in \B_t$, the current policy $\pi_{\theta_t}$ generates a group of $G$ responses $\{o_g\}_{g=1}^G$.
A verifier then assigns a binary reward to each response, yielding the reward set $R=\{r_g\;|\;r_g\in\{0,1\}\}_{g=1}^G$, where $r_g=r(\x^{\mathrm{S}},o_g,a^{\mathrm{S}})$.
Then, GRPO forms a group-relative advantage by normalizing rewards within the group:
\begin{equation*}
\label{eqn:advantage}
A_{g} = \frac{r_g - \mathrm{mean} (R)}{\mathrm{std} (R) + \varepsilon},
\end{equation*}
where $\varepsilon>0$ is a constant, $\mathrm{mean}(R)$ and $\mathrm{std}(R)$ denote the mean and standard deviation of $R$.

Following~\cite{yu2025dapo}, we use the token-level GRPO loss and omit the Kullback-Leibler~{(KL)} penalty term for each $(\x^{\mathrm{S}},a^{\mathrm{S}})$, i.e.,

\begin{equation}
\label{eqn:grpo_loss}
\begin{aligned}
\ell\,&(\theta_t,\theta_{t-1};\x^{\mathrm{S}},a^{\mathrm{S}}) = -\E_{\{o_g\}_{g=1}^G\sim \pi_{\theta_{t-1}}(\cdot\mid \x^{\mathrm{S}})} \frac{1}{\sum_{g=1}^{G}\lvert o_g\rvert} \sum_{g=1}^{G}\sum_{j=1}^{\lvert o_g\rvert}\min\!\left(\rho_{g,j}A_g,\,\bar\rho_{g,j}A_g\right),
\end{aligned}
\end{equation}
where $\bar \rho_{g,j}=\mathrm{clip}\!\left(\rho_{g,j},\,1-\epsilon_{\mathrm{low}},\,1+\epsilon_{\mathrm{high}}\right)$, with $\mathrm{clip}(u,a,b)=\min\{\max\{u,a\},\,b\}$, and $\rho_{g,j} = \pi_{\theta_t}\!\big(o_{g,j}\mid \x,\,o_{g,<j}\big) / \pi_{\theta_{t-1}}\!\big(o_{g,j}\mid \x,\,o_{g,<j}\big)$.

\textbf{Mitigating advantage collapse.}
In GRPO, the group-normalized advantages can collapse to zero when all responses in a group receive the same verifier reward, thereby eliminating the policy gradients.
To mitigate this, \cite{yu2025dapo} proposed dynamic sampling that filters out prompts with identical group rewards and resamples to fill the training batch, while \cite{zheng2025act} proposed a more sample-efficient method that maintains, for each $\x_i^{\mathrm{S}}$, a history of its rollout rewards.
Let $n_i(t)$ denote the number of times $\x_i^{\mathrm{S}}$ has been sampled for rollout before round~$t$. Its history is
\begin{equation*}
\tau_{i,t} = \bigl((e_{i,1},R_{i,1}),\ldots,(e_{i,n_i(t)},R_{i,n_i(t)})\bigr),
\end{equation*}
where $e_{i,s}$ with $s\in\{1,\ldots,n_i(t)\}$ is the index of the training round in which $\x_i^{\mathrm{S}}$ is sampled for rollout for the $s$-th time, and $R_{i,s}$ denotes the set of verifier rewards from the responses generated on that occasion.
An indicator $I_{i,s}$ and an online counter $z_{i,t}$ are then defined for each prompt $\x_i^{\mathrm{S}}$ as
\begin{equation*}
I_{i,s} = \begin{cases}
1, & \text{if all rewards in }R_{i,s}\text{ are identical},\\
0, & \text{otherwise},
\end{cases}
\end{equation*}
and
\begin{equation*}
z_{i,t} = \max\left\{ k\in\{0,\ldots,n_i(t)\}\ \middle|\ \prod_{s=n_i(t)-k+1}^{n_i(t)} I_{i,s}=1 \right\},
\end{equation*}
where the empty product is defined as $1$, so that $z_{i,t}=0$ for prompts with no previous rollouts.
The counter measures the number of most recent consecutive sampling events for $\x_i^{\mathrm{S}}$ that yield zero GRPO advantages in Eq.~\eqref{eqn:grpo_loss}.

Following the filtering probability in \cite{zheng2025act}, we assign each prompt $\x_i^{\mathrm{S}}$ a rollout-waste penalty that increases with persistently zero advantages:
\begin{equation}
\label{eqn:adv_vanish_panish}
1-p_e^{\,z_{i,t}},
\end{equation}
where $p_e\in(0,1)$ controls how rapidly the penalty increases with zero-advantage rollouts.


\section{Learning-to-Reweight Rollout (L2RR)}
\label{sec:approach}

In this section, we present our proposed algorithm and theoretical analysis.

\textbf{Formulation.} Due to source-target datasets mismatch, a policy that is optimal for $\D_{\rmsrc}$ need not yield a policy that performs best on $\D_{\rmtgt}$.
Inspired by importance reweighting for mitigating dataset mismatch~\citep{sugiyama2007covariate,ren2018learning}, we adapt this principle to RLVR to jointly improve target-domain generalization and rollout sample efficiency.
Specifically, we maintain a time-varying sampling distribution $\q_t\in\Delta^{N_S}$ for $t\in[T]$, where $q_{t,i}>0$ and $\sum_{i=1}^{N_S} q_{t,i}=1$.
At each round, we optimize the
\begin{equation*}
\begin{aligned}
L_{\mathrm{GRPO}}(\q_t) = \E_{(\x^{\mathrm{S}},a^{\mathrm{S}})\sim\q_t} \, \ell\,(\theta_t,\theta_{t-1};\x^{\mathrm{S}},a^{\mathrm{S}}).
\end{aligned}
\end{equation*}

The target owner evaluates the policy locally on the held-out target validation set, with the target objective defined as:
\begin{equation*}
\label{eqn:target_loss}
L_{\rmtgt}(\theta_t) = -\E_{\x\sim \D_{\rmtgt}}\E_{y\sim \pi_{\theta_t}(\cdot\mid \x)}\big[r(\x,y)\big].
\end{equation*}
Here, $D_{\mathrm{tgt}}$ is used to define the target-side local data, but its prompts are not exposed to the trainer.

This leads to the following objective function:
\begin{equation}
\label{eqn:q_GRPO}
\begin{aligned}
\min_{\q_1,\ldots,\q_T\in\Delta^{N_S}}\quad
& \E\!\left[L_{\rmtgt}(\theta_{T+1})\right]\\
\text{s.t.}\quad
& \B_t\sim\q_t, \quad \theta_{t+1} = \theta_t-\frac{\eta_\theta}{B} \sum_{i\in\B_t}g_{t,i}, \qquad t=1,\ldots,T,
\end{aligned}
\end{equation}
where $\theta_1$ is the initial policy parameter, $g_{t,i}$ is the per-example GRPO gradient, and each $\q_t$ is chosen using the training history available before round $t$.
The expectation is over prompt sampling and rollout generation.

Although we can upweight the sampling distribution on prompts whose rollouts most improve target held-out performance, those source prompts can still be either beyond the model's current capability or already solved so consistently that their GRPO advantages become degenerate.
In both cases, repeatedly sampling them wastes rollout budget.
To improve the sample efficiency, we use the rollout-waste penalty in Eq.~\eqref{eqn:adv_vanish_panish}, which penalizes the prompts that consecutively produce zero GRPO advantages.
We quantify this history dependent rollout waste along this sampling rule by
\begin{equation*}
L_{\mathrm{ws}}(\q_{1:T}) = \E\!\left[
\frac{1}{TB} \sum_{t=1}^{T}\sum_{b=1}^{B} \Bigl(1-p_e^{\,z_{i_{t,b},t}} \Bigr) \right],
\end{equation*}
where $\q_{1:T}=(\q_1,\ldots,\q_T)$ and $i_{t,b}$ is the source-prompt index of the $b$-th sample in $\B_t$.
Minimizing $L_{\mathrm{ws}}$ makes the sampler place less mass on prompts with large $z_{i,t}$, reducing the cost spent on uninformative rollouts.

Therefore, our learning objective is as follows:
\begin{equation}
\label{eqn:main_obj}
\begin{aligned}
\min_{\q_1,\ldots,\q_T\in\Delta^{N_S}} \quad
& \E\!\left[L_{\rmtgt}(\theta_{T+1})\right] + \lambda L_{\mathrm{ws}}(\q_{1:T})\\
\text{s.t.}\quad
& \B_t\sim\q_t, \quad \theta_{t+1} = \theta_t-\frac{\eta_\theta}{B} \sum_{i\in\B_t}g_{t,i}, \qquad t=1,\ldots,T,
\end{aligned}
\end{equation}
where $\lambda\ge0$ is the regularization hyperparameter.

\textbf{Bi-level optimization task.}
The formulation in Eq.~\eqref{eqn:main_obj} motivates a bi-level optimization surrogate~\citep{colson2007overview}.
At iteration $t$, we first update the policy by one GRPO step on the batch $\B_t$ sampled from $\q_t$; following GRESO~\citep{zheng2025act}, prompts are drawn from $\q_t$ in sampling rounds (Table~\ref{tab:hyperparams}) until $B$ prompts with non-degenerate rewards are collected, and all generated responses count toward the rollout cost:
\begin{equation}
\label{eqn:our_grpo}
\theta_{t+1} = \theta_t -\eta_\theta\cdot \frac{1}{B}\sum_{i\in\B_t} g_{t,i},
\end{equation}
where $g_{t,i}$ is the per-example GRPO gradient computed from the rollouts of prompt $\x_i^{\mathrm{S}}$ under $\pi_{\theta_t}$.

After updating the policy, we update the sampling distribution $\q_t$.
Since the dependence of $L_{\rmtgt}(\theta_{t+1})$ on $\q_t$ is implicit through discrete sampling, following the one-step unrolling surrogate used in importance weighting~\citep{ren2018learning,shu2019meta},
we form a differentiable look-ahead update on $\B_t$ with $\theta_{t+1}(\q_t)=\theta_t-\eta_\theta\sum_{i\in\B_t}\bar q_{t,i}\,g_{t,i}$, where $\bar q_{t,i}=q_{t,i}/\sum_{j\in\B_t}q_{t,j}$.
Let $g^{\rmtgt}_{t+1}=\nabla_\theta L_{\rmtgt}(\theta)\big|_{\theta=\theta_{t+1}(\q_t)}$ and $\bar g_t=\sum_{i\in\B_t}\bar q_{t,i}\,g_{t,i}$.
Treating $\B_t$, $\theta_t$, and the per-example gradients as fixed, the one-step surrogate yields, for each $i\in\B_t$,
\begin{equation}
\label{eqn:grad_tgt}
\widehat g^{\rmtgt}_{t,i} = \frac{\partial}{\partial q_{t,i}}L_{\rmtgt}\!\big(\theta_{t+1}(\q_t)\big) = -\eta_\theta \frac{\left\langle g^{\rmtgt}_{t+1},\,g_{t,i}-\bar g_t\right\rangle}{\sum_{j\in\B_t}q_{t,j}}.
\end{equation}
The directional derivatives $\langle g^{\rmtgt}_{t+1},g_{t,i}-\bar g_t\rangle$ can be computed by the target owner without sharing target prompts.
We set $\widehat g^{\rmtgt}_{t,i}=0$ for $i\notin\B_t$.
This interface provides target-data locality and restricted disclosure, while establishing formal privacy guarantees is left to future work.

For the vanishing-advantage regularizer, we use the online surrogate $L_{\mathrm{ws}}(\q_t)=\sum_{i=1}^{N_S} q_{t,i}\big(1-p_e^{\,z_{i,t}}\big)$, whose coordinate gradient is $\widehat g^{\mathrm{ws}}_{t,i} = \lambda\big(1-p_e^{\,z_{i,t}}\big)$ for $i=1,\ldots,N_S$.

Combining $\widehat g_{t,i}=\widehat g^{\rmtgt}_{t,i}+\widehat g^{\mathrm{ws}}_{t,i}$, we update $\q_t$ via an exponentiated-gradient step:
\begin{equation}
\label{eq:q_update_impl}
q_{t+1,i} = \frac{q_{t,i}\exp\!\big(-\eta_q\,\widehat g_{t,i}\big)}{\sum_{j=1}^{N_S}q_{t,j}\exp\!\big(-\eta_q\,\widehat g_{t,j}\big)}.
\end{equation}

\textbf{From sampler optimization to rollout allocation.}
The sampler update relies on a target-guided gradient estimate, whose accuracy depends on the gradients $\{g_{t,i}\}_{\x_i\in\B_t}$ and is thus affected by the per-prompt rollout count $G_{i,t}$.
In Section~\ref{sec:theorem}, our theoretical analysis makes this dependence explicit: lowering the variance of the target-guided gradient estimator tightens the target-loss bound, with other terms held fixed.
Therefore, \emph{the bound motivates our rollout allocation}: with a given per-round rollout budget, we allocate more rollouts to prompts that exert larger influence on the target-guided sampler gradient update, and fewer rollouts to prompts that repeatedly yield degenerate GRPO advantages.

We first define the time-varying rollout budget based on the history-dependent informativeness scores.
For each prompt $\x_i\in\B_t$, let $s_{i,t}=p_e^{\,z_{i,t}}$, and set $\bar s_t = \frac{1}{B}\sum_{i\in\B_t} s_{i,t}$.
The rollout budget is defined as $\mathrm{RO}_t = BG\cdot \mathrm{clip}\big(\bar s_t,\rho_{\min},1\big)$, where $\rho_{\min}\in[G_{\min}/G,1]$ enforces a minimum rollout budget.

Then, at iteration $t$, we use the rollout budget $\mathrm{RO}_t$ and optimize $\{G_{i,t}\}_{i\in\B_t}$ to reduce the estimation error of the outer-gradient used to update $\q_t$.
Assume that the variance of the per-prompt outer-gradient contribution decays inversely with the group size, e.g., $\mathrm{Var}\!\left(\widehat g^{\rmtgt}_{t,i}\right) \le \frac{c_{i,t}}{G_{i,t}}$, where $c_{i,t}\ge 0$ is an iteration-dependent variance proxy for rollout allocation.
Under this model, we use a weighted allocation proxy to allocate more rollouts to prompts that contribute more to the sampler update and are not persistently degenerate.
Specifically, we assign each prompt a weight $w_{i,t}=p_e^{\,z_{i,t}}\,u_{i,t}$, $u_{i,t}=\Big|\big\langle g^{\rmtgt}_{t+1},\,g_{t,i}-\bar g_t\big\rangle\Big|$, where $u_{i,t}$ measures the magnitude of the contribution of prompt $\x_i^{\mathrm{S}}$ to the target-guided sampler gradient, and $p_e^{\,z_{i,t}}$ downweights prompts with large waste counters.

This leads to the following rollout budget allocation problem:
\begin{equation*}
\begin{aligned}
\label{eq:budget_alloc_problem}
& \quad\quad\quad \min_{\{G_{i,t}\}}
\ \sum_{i\in\B_t}\frac{w_{i,t}\,c_{i,t}}{G_{i,t}}\\
& \text{s.t.}\quad \sum_{i\in\B_t} G_{i,t} \le \mathrm{RO}_t,\;\; G_{\min}\le G_{i,t}\le G.
\end{aligned}
\end{equation*}
For positive weights and proxies, the continuous optimum satisfies $G_{i,t}\propto\sqrt{w_{i,t}c_{i,t}}$ when no bounds are active.
In our implementation, we set the proxy to $c_{i,t}=w_{i,t}^2$, so that $\sqrt{w_{i,t}c_{i,t}}=w_{i,t}^{3/2}$.
Letting $\beta_{i,t}=\sqrt{w_{i,t}\,c_{i,t}}$ and normalizing within $\B_t$ as $\bar\beta_{i,t}=\beta_{i,t}/\sum_{j\in\B_t}\beta_{j,t}$, we allocate
\begin{equation}
\label{eqn:rollouts}
G_{i,t} = \mathrm{clip}\!\left( \left\lfloor \mathrm{RO}_t\cdot \bar\beta_{i,t}\right\rfloor,\; G_{\min},\; G \right).
\end{equation}

This allocation aims to reduce the variance of the target-guided gradient estimator used in the $\q_t$ update, motivated by the noise dependence of our regret bound in Theorem~\ref{thm:informal_main}.
We summarize our method in Algorithm~\ref{alg:implicit_hyper_q_grpo}.

\begin{algorithm}[t]
\caption{Learning-to-Reweight Rollouts (L2RR)}
\label{alg:implicit_hyper_q_grpo}
\begin{algorithmic}[1]
\REQUIRE Source dataset $\D_{\rmsrc}$, target-side feedback oracle $\mathcal{O}_{\mathrm{tgt}}$, weight $\lambda$, iterations $T$, batch size $B$, group size threshold $G_{\min}$ and G, learning rate $\eta_\theta$ and $\eta_q$, policy $\pi_\theta$, verifier $r$
\vspace{2mm}
\STATE \textbf{Initialize} $\q_1$ as uniform distribution, $\theta_1$, and $z_{i,1}=0$.
\FOR{$t=1,\ldots,T$}
\STATE Sample mini-batch $\B_t=\{\x^{\mathrm{S}}_{i_1},\ldots,\x^{\mathrm{S}}_{i_B}\}\sim \q_t$
\STATE Set the rollout number $G_{i,t}$ according to Eq.~\eqref{eqn:rollouts}
\STATE Sample $G_{i,t}$ rollouts $\{o_{i,g}\}_{g=1}^{G_{i,t}}$ for $\x^{\mathrm{S}}_i\in\B_t$ and compute rewards $r_{i,g}=r(\x^{\mathrm{S}}_i,o_{i,g},a^{\mathrm{S}})$
\STATE Update policy $\theta_{t+1}$ according to Eq.~\eqref{eqn:our_grpo}
\STATE Update sampler $\q_{t+1}$ according to Eq.~\eqref{eq:q_update_impl}
\ENDFOR
\STATE \textbf{Output} policy $\pi_{\theta_{T+1}}$.
\end{algorithmic}
\end{algorithm}

\section{Theoretical Analysis}
\label{sec:theorem}

In this section, we provide a standard online learning analysis for the sampler update in L2RR. The analysis views the sampling distribution update as an exponentiated-gradient step on the simplex and derives a sublinear regret bound for the linear losses induced by its feedback~\citep{hazan2022online}. We also quantify the effect of the rollout-waste penalty and distinguish this regret guarantee from the conditional target-loss analysis in \textbf{Appendix~\ref{app:proofs}}.

Let $\widehat\ell_t(\q)=\langle\widehat g_t,\q\rangle$, where $\widehat g_t$ is the feedback in Eq.~\eqref{eq:q_update_impl}. We compare the sampler with the best fixed distribution on the same realized loss sequence:
\begin{equation}
\label{eq:linear_regret}
\widehat R_T=\frac{1}{T}\left(\sum_{t=1}^T\widehat\ell_t(\q_t)-\min_{\q\in\Delta^{N_S}}\sum_{t=1}^T\widehat\ell_t(\q)\right).
\end{equation}

\begin{myThm}[Sampler regret]
\label{thm:informal_main}
Initialize $\q_1$ uniformly and assume $T^{-1}\sum_{t=1}^T\E\|\widehat g_t\|_\infty^2\le H^2$, with $H$ independent of $T$. For any fixed $a>0$, use $\eta_q=a/\sqrt T$ throughout the run. Then
\begin{equation}
\label{eq:sampler_regret_main}
\E\left[\widehat R_T\right]
\le\frac{\log N_S/a+aH^2/2}{\sqrt T}.
\end{equation}
Let $\widehat R_T^{\rmtgt}$ denote the same regret with $\widehat\ell_t^{\rmtgt}(\q)=\langle\widehat g_t^{\rmtgt},\q\rangle$. For a fixed per-run weight $\lambda_T\ge0$,
\begin{equation}
\label{eq:target_feedback_regret_main}
\E\left[\widehat R_T^{\rmtgt}\right]
\le\frac{\log N_S/a+aH^2/2}{\sqrt T}+\lambda_T.
\end{equation}
\end{myThm}

Thus the cumulative feedback regret is $\O(\sqrt T)$ and its average is $\O(T^{-1/2})$. Choosing $\lambda_T\le c_\lambda/\sqrt T$, held fixed within each run, gives the same rates for the target-feedback losses. The second-moment term includes gradient-estimation noise, connecting the bound to rollout allocation without assuming that the weighted proxy minimizes that noise.

\section{Experiments}
\label{sec:experiments}

In this section, we evaluate the proposed algorithm on multiple benchmarks with three different models.
We also conduct ablation studies on our approach.

\subsection{Experimental Setup}
\label{sec:experimental_setup}

\textbf{Models.}
We test L2RR on Qwen2.5-Math-1.5B~\citep{yang2024qwen2}, DeepSeek-R1-Distill-Qwen-1.5B~\citep{guo2025deepseek}, and Qwen2.5-Math-7B~\citep{yang2024qwen2}.
For Qwen2.5-Math-1.5B/7B models, we use 4096 as the context length, as it is the maximum context length for those two models.
For DeepSeek-R1-Distill-Qwen-1.5B, we set the context length to 8192 (2048 prompt and 6144 response tokens), following the experimental settings in the work of~\cite{zheng2025act}.

\textbf{Datasets.} For source datasets, we consider: I). DAPO+MATH: we combine the DAPO dataset~\citep{yu2025dapo}, which contains only integer solutions, with the MATH dataset~\citep{hendrycks2measuring}, which contains LaTeX-formatted solutions.
II). OPEN-R1 30k subset (R1): A 30,000-example subset of the OPEN-R1 math dataset\footnote{\url{https://huggingface.co/datasets/open-r1/OpenR1-Math-220k}}.

For target datasets, we use six widely used mathematical reasoning benchmarks to evaluate the performance: Math500~\citep{hendrycks2measuring}, AIME24\footnote{\url{https://huggingface.co/datasets/Maxwell-Jia/AIME_2024}}, AMC\footnote{\url{https://huggingface.co/datasets/AI-MO/aimo-validation-amc}}, Minerva Math~\citep{lewkowycz2022solving}, Gaokao~\citep{zhang2023evaluating}, Olympiad Bench~\citep{he2024olympiadbench}.

\textbf{Setups.} Following~\citep{zheng2025act}, we implement L2RR using verl~\citep{sheng2025hybridflow} and vLLM~\citep{kwon2023efficient} for rollouts.
We use Math-Verify\footnote{\url{https://github.com/huggingface/Math-Verify}} to verify the generated answers.
Hyperparameters are listed in \textbf{Appendix~\ref{app:exp_details}}.
Each target benchmark is partitioned into fixed, disjoint 10\% validation and 90\% test splits shared across all methods and five training seeds.
L2RR uses validation data for sampler learning; all methods use it for checkpoint selection.
Following~\citep{wang2025reinforcement,zheng2025act}, we evaluate checkpoints on the validation splits every 50 steps and report test performance for the checkpoint with the best macro-average validation accuracy across the six benchmarks.

We also evaluate source-only baselines without target-validation-based checkpoint selection and Target-Mix baselines that directly incorporate target validation examples into policy training, requiring target-data access beyond our setting.
Results are reported in \textbf{Appendix~\ref{app:emprical_fair}}.

\textbf{Contenders.} We compare the proposed algorithm with three strong contenders that consider rollout efficiency.
The decoupled clip and dynamic sampling policy optimization~{(\textsc{DAPO})} mitigates advantage collapse by dynamically resampling prompts to ensure non-degenerate verifier-reward variance in each GRPO group~\citep{yu2025dapo}.
GRPO with efficient selective rollout~{(\textsc{GRESO})} reduces rollout cost via online pre-rollout filtering that predicts and skips prompts likely to produce degenerate near-zero GRPO updates~\citep{zheng2025act}.
Adaptive rollout and response reuse policy optimization~{(\textsc{AR3PO})} improves sampling efficiency by adaptive rollout allocation and response reuse, with mechanisms to control the mismatch between reused responses and the current policy~\citep{zhang2025improving}.

\begin{table*}[t]
\centering
\caption{\textbf{Performance (\%) comparison across six math reasoning benchmarks.}
We train three models on source datasets DAPO+MATH (DM) and the Open R1 subset (OR1) and test on six benchmarks.
\# Rollout denotes cumulative source responses, in millions.
RO ratio is the DAPO count divided by the method count. Accuracies are mean $\pm$ standard deviation over five training seeds.}
\label{tab:perf_rollouts}
\setlength{\tabcolsep}{6pt}
\renewcommand{\arraystretch}{1.15}
\resizebox{0.98\textwidth}{!}{
\begin{tabular}{c|c|cccccc|c|c|c}
\hline
Dataset & Method & Math500 & AIME24 & AMC & Gaokao & Miner. & Olymp. & Avg. & \# Rollout & RO ratio \\
\hline\hline

\multicolumn{11}{c}{\textit{Qwen2.5-Math-1.5B}}\\
\hline
\multirow{4}{*}{DM}
  & DAPO
  & {{76.3}}{\tiny {$\pm$ 1.8}} & {{15.2}}{\tiny {$\pm$ 3.1}} & {{60.9}}{\tiny {$\pm$ 2.7}}
  & {{63.8}}{\tiny {$\pm$ 2.3}} & {{31.6}}{\tiny {$\pm$ 2.5}} & {{38.2}}{\tiny {$\pm$ 1.2}}
  & {{47.7}}{\tiny {$\pm$ 1.5}} & 7.6M & 1.0$\times$ \\
  & GRESO
  & {{75.8}}{\tiny {$\pm$ 1.8}} & {{14.1}}{\tiny {$\pm$ 3.0}} & {{60.8}}{\tiny {$\pm$ 2.8}}
  & {{65.5}}{\tiny {$\pm$ 2.1}} & {{32.8}}{\tiny {$\pm$ 2.8}} & {{38.0}}{\tiny {$\pm$ 1.5}}
  & {{47.8}}{\tiny {$\pm$ 1.6}} & 3.3M & 2.3$\times$ \\
  & AR3PO
  & {{77.6}}{\tiny {$\pm$ 1.3}} & {{15.7}}{\tiny {$\pm$ 3.1}} & {{60.8}}{\tiny {$\pm$ 2.5}}
  & {{65.3}}{\tiny {$\pm$ 2.3}} & {\textbf{{35.4}}}{\tiny {$\pm$ 2.5}} & {{38.6}}{\tiny {$\pm$ 1.9}}
  & {{48.9}}{\tiny {$\pm$ 1.5}} & 2.5M & 3.0$\times$ \\
  & L2RR
  & {\textbf{{79.4}}}{\tiny {$\pm$ 1.7}} & {\textbf{{17.2}}}{\tiny {$\pm$ 2.4}} & {\textbf{{61.8}}}{\tiny {$\pm$ 2.1}}
  & {\textbf{{66.8}}}{\tiny {$\pm$ 2.1}} & {{35.2}}{\tiny {$\pm$ 2.1}} & {\textbf{{41.4}}}{\tiny {$\pm$ 0.9}}
  & {\textbf{{50.3}}}{\tiny {$\pm$ 1.3}} & 3.0M & {2.5$\times$} \\
\cline{1-11}
\multirow{4}{*}{OR1}
  & DAPO
  & {{76.5}}{\tiny {$\pm$ 1.9}} & {{15.7}}{\tiny {$\pm$ 2.7}} & {{49.5}}{\tiny {$\pm$ 2.8}}
  & {{65.1}}{\tiny {$\pm$ 2.3}} & {{30.2}}{\tiny {$\pm$ 2.6}} & {{39.2}}{\tiny {$\pm$ 1.7}}
  & {{46.0}}{\tiny {$\pm$ 1.6}} & 3.8M & 1.0$\times$ \\
  & GRESO
  & {{75.8}}{\tiny {$\pm$ 2.1}} & {{18.5}}{\tiny {$\pm$ 2.8}} & {{49.7}}{\tiny {$\pm$ 2.5}}
  & {{64.9}}{\tiny {$\pm$ 2.1}} & {{29.5}}{\tiny {$\pm$ 2.5}} & {{39.0}}{\tiny {$\pm$ 0.8}}
  & {{46.2}}{\tiny {$\pm$ 1.5}} & 1.6M & 2.4$\times$ \\
  & AR3PO
  & {{76.2}}{\tiny {$\pm$ 1.8}} & {{18.3}}{\tiny {$\pm$ 3.2}} & {{49.8}}{\tiny {$\pm$ 2.5}}
  & {{64.7}}{\tiny {$\pm$ 2.1}} & {{29.7}}{\tiny {$\pm$ 2.5}} & {{39.2}}{\tiny {$\pm$ 1.1}}
  & {{46.3}}{\tiny {$\pm$ 1.5}} & 1.4M & {2.7$\times$} \\
  & L2RR
  & {\textbf{{78.3}}}{\tiny {$\pm$ 1.5}} & {\textbf{{19.8}}}{\tiny {$\pm$ 2.7}} & {\textbf{{51.1}}}{\tiny {$\pm$ 2.3}}
  & {\textbf{{66.2}}}{\tiny {$\pm$ 1.7}} & {\textbf{{30.6}}}{\tiny {$\pm$ 1.8}} & {\textbf{{41.5}}}{\tiny {$\pm$ 1.1}}
  & {\textbf{{47.9}}}{\tiny {$\pm$ 1.3}} & 1.4M & {2.7$\times$} \\
\hline\hline

\multicolumn{11}{c}{\textit{DeepSeek-R1-Distill-Qwen-1.5B}}\\
\hline
\multirow{4}{*}{DM}
  & DAPO
  & {{85.9}}{\tiny {$\pm$ 2.1}} & {{34.8}}{\tiny {$\pm$ 3.3}} & {{70.5}}{\tiny {$\pm$ 2.9}}
  & {{78.1}}{\tiny {$\pm$ 2.3}} & {{34.4}}{\tiny {$\pm$ 2.8}} & {{54.3}}{\tiny {$\pm$ 1.8}}
  & {{59.7}}{\tiny {$\pm$ 1.7}} & 2.4M & 1.0$\times$ \\
  & GRESO
  & {{85.7}}{\tiny {$\pm$ 2.3}} & {{34.3}}{\tiny {$\pm$ 3.1}} & {{70.1}}{\tiny {$\pm$ 3.1}}
  & {{77.5}}{\tiny {$\pm$ 2.6}} & {{33.1}}{\tiny {$\pm$ 2.6}} & {{53.1}}{\tiny {$\pm$ 2.3}}
  & {{59.0}}{\tiny {$\pm$ 1.8}} & 1.6M & 1.5$\times$ \\
  & AR3PO
  & {{86.4}}{\tiny {$\pm$ 1.9}} & {{34.6}}{\tiny {$\pm$ 3.6}} & {{70.2}}{\tiny {$\pm$ 3.5}}
  & {{77.9}}{\tiny {$\pm$ 2.4}} & {{33.9}}{\tiny {$\pm$ 2.6}} & {{53.9}}{\tiny {$\pm$ 1.8}}
  & {{59.5}}{\tiny {$\pm$ 1.8}} & 1.7M & 1.4$\times$ \\
  & L2RR
  & {\textbf{{88.2}}}{\tiny {$\pm$ 1.7}} & {\textbf{{36.7}}}{\tiny {$\pm$ 2.9}} & {\textbf{{71.8}}}{\tiny {$\pm$ 2.5}}
  & {\textbf{{79.5}}}{\tiny {$\pm$ 2.3}} & {\textbf{{35.3}}}{\tiny {$\pm$ 2.3}} & {\textbf{{56.8}}}{\tiny {$\pm$ 1.7}}
  & {\textbf{{61.4}}}{\tiny {$\pm$ 1.5}} & 1.6M & 1.5$\times$ \\
\cline{1-11}
\multirow{4}{*}{OR1}
  & DAPO
  & {84.8}{\tiny {$\pm$ 2.5}} & {{22.2}}{\tiny {$\pm$ 2.9}} & {\textbf{{68.1}}}{\tiny {$\pm$ 2.7}}
  & {{73.2}}{\tiny {$\pm$ 2.7}} & {{33.4}}{\tiny {$\pm$ 2.9}} & {{52.4}}{\tiny {$\pm$ 1.9}}
  & {{55.7}}{\tiny {$\pm$ 1.8}} & 2.4M & 1.0$\times$ \\
  & GRESO
  & {{85.6}}{\tiny {$\pm$ 2.3}} & {{24.1}}{\tiny {$\pm$ 3.3}} & {{66.3}}{\tiny {$\pm$ 2.8}}
  & {{74.4}}{\tiny {$\pm$ 2.3}} & {{32.9}}{\tiny {$\pm$ 2.5}} & {{54.3}}{\tiny {$\pm$ 1.9}}
  & {{56.3}}{\tiny {$\pm$ 1.7}} & 1.5M & 1.6$\times$ \\
  & AR3PO
  & {{85.1}}{\tiny {$\pm$ 2.2}} & {{23.7}}{\tiny {$\pm$ 3.5}} & {{66.8}}{\tiny {$\pm$ 2.5}}
  & {{74.2}}{\tiny {$\pm$ 2.2}} & {{33.0}}{\tiny {$\pm$ 2.5}} & {{53.8}}{\tiny {$\pm$ 1.9}}
  & {{56.1}}{\tiny {$\pm$ 1.7}} & 1.6M & 1.5$\times$ \\
  & L2RR
  & {\textbf{{86.7}}}{\tiny {$\pm$ 1.8}} & {\textbf{{25.9}}}{\tiny {$\pm$ 2.8}} & {{67.7}}{\tiny {$\pm$ 2.6}}
  & {\textbf{{76.1}}}{\tiny {$\pm$ 2.1}} & {\textbf{{34.8}}}{\tiny {$\pm$ 2.3}} & {\textbf{{57.4}}}{\tiny {$\pm$ 1.4}}
  & {\textbf{{58.1}}}{\tiny {$\pm$ 1.5}} & 1.5M & 1.6$\times$ \\
\hline\hline

\multicolumn{11}{c}{\textit{Qwen2.5-Math-7B}}\\
\hline
\multirow{4}{*}{DM}
  & DAPO
  & {{82.3}}{\tiny {$\pm$ 2.1}} & {\textbf{{33.3}}}{\tiny {$\pm$ 3.1}} & {{77.8}}{\tiny {$\pm$ 2.5}}
  & {{72.3}}{\tiny {$\pm$ 2.1}} & {{34.9}}{\tiny {$\pm$ 2.3}} & {{45.1}}{\tiny {$\pm$ 1.7}}
  & {{57.6}}{\tiny {$\pm$ 1.6}} & 13.1M & 1.0$\times$ \\
  & GRESO
  & {{81.4}}{\tiny {$\pm$ 2.1}} & {{30.6}}{\tiny {$\pm$ 3.3}} & {{79.5}}{\tiny {$\pm$ 2.5}}
  & {{70.5}}{\tiny {$\pm$ 2.3}} & {{34.6}}{\tiny {$\pm$ 2.4}} & {{45.4}}{\tiny {$\pm$ 1.9}}
  & {{57.0}}{\tiny {$\pm$ 1.6}} & 6.3M & 2.1$\times$ \\
  & AR3PO
  & {{82.1}}{\tiny {$\pm$ 2.2}} & {{31.5}}{\tiny {$\pm$ 2.8}} & {{78.5}}{\tiny {$\pm$ 3.3}}
  & {{71.9}}{\tiny {$\pm$ 2.3}} & {{35.1}}{\tiny {$\pm$ 2.7}} & {{44.7}}{\tiny {$\pm$ 1.9}}
  & {{57.3}}{\tiny {$\pm$ 1.7}} & 6.6M & 2.0$\times$ \\
  & L2RR
  & {\textbf{{83.5}}}{\tiny {$\pm$ 2.1}} & {{32.4}}{\tiny {$\pm$ 2.7}} & {\textbf{{80.4}}}{\tiny {$\pm$ 2.6}}
  & {\textbf{{72.8}}}{\tiny {$\pm$ 2.2}} & {\textbf{{36.3}}}{\tiny {$\pm$ 2.4}} & {\textbf{{47.5}}}{\tiny {$\pm$ 1.9}}
  & {\textbf{{58.8}}}{\tiny {$\pm$ 1.6}} & 6.0M & 2.2$\times$ \\
\cline{1-11}
\multirow{4}{*}{OR1}
  & DAPO
  & {{82.2}}{\tiny {$\pm$ 1.9}} & {{32.0}}{\tiny {$\pm$ 2.7}} & {{64.3}}{\tiny {$\pm$ 2.5}}
  & {{68.1}}{\tiny {$\pm$ 2.5}} & {{34.7}}{\tiny {$\pm$ 2.3}} & {{47.5}}{\tiny {$\pm$ 1.8}}
  & {54.8}{\tiny {$\pm$ 1.5}} & 11.4M & 1.0$\times$ \\
  & GRESO
  & {{81.7}}{\tiny {$\pm$ 2.3}} & {{33.1}}{\tiny {$\pm$ 3.0}} & {{65.8}}{\tiny {$\pm$ 2.4}}
  & {{68.0}}{\tiny {$\pm$ 2.4}} & {{35.3}}{\tiny {$\pm$ 2.5}} & {{47.3}}{\tiny {$\pm$ 2.1}}
  & {{55.2}}{\tiny {$\pm$ 1.7}} & 3.4M & 3.4$\times$ \\
  & AR3PO
  & {{82.0}}{\tiny {$\pm$ 2.6}} & {{33.5}}{\tiny {$\pm$ 3.1}} & {{65.9}}{\tiny {$\pm$ 2.8}}
  & {{67.5}}{\tiny {$\pm$ 2.8}} & {{36.5}}{\tiny {$\pm$ 2.8}} & {{46.9}}{\tiny {$\pm$ 1.8}}
  & {{55.4}}{\tiny {$\pm$ 1.8}} & 3.6M & 3.2$\times$ \\
  & L2RR
  & {\textbf{{83.1}}}{\tiny {$\pm$ 2.1}} & {\textbf{{35.0}}}{\tiny {$\pm$ 2.6}} & {\textbf{{67.2}}}{\tiny {$\pm$ 2.3}}
  & {\textbf{{69.3}}}{\tiny {$\pm$ 2.3}} & {\textbf{{37.2}}}{\tiny {$\pm$ 2.3}} & {\textbf{{48.9}}}{\tiny {$\pm$ 1.7}}
  & {\textbf{56.8}}{\tiny {$\pm$ 1.5}} & 3.3M & 3.5$\times$ \\
\hline
\end{tabular}
}
\vspace{-4mm}
\end{table*}

\subsection{Main Results}

Table~\ref{tab:perf_rollouts} reports the comparison results of the proposed algorithm with three state-of-the-art contenders across six math reasoning benchmarks, three models spanning two parameter scales, and two source training sets.
We report the checkpoint with the best macro-average validation accuracy, following the setting in previous work~\citep{zheng2025act,zhang2025improving}.
Across all settings, L2RR achieves the highest average performance while consistently reducing the rollout cost relative to DAPO.
On Qwen2.5-Math-1.5B trained on DM, L2RR improves the average from $47.7\%$ to $50.3\%$ while reducing rollouts from $7.6$M to $3.0$M.
These results suggest that leveraging limited target-side feedback to learn the source-prompt sampling distribution produces more cost-effective rollouts and improves generalization to the target domain.

GRESO and AR3PO reduce rollouts relative to DAPO by exploiting only source-side rollout statistics, with averages close to DAPO in most configurations.
In contrast, L2RR more consistently improves average accuracy while maintaining comparable rollout budgets to these baselines, by exploiting a few target domain information.
For example, on Qwen2.5-Math-7B trained on DM, L2RR reaches $58.8\%$ average with $6.0$M rollouts, outperforming GRESO and AR3PO.
Overall, L2RR achieves a stronger accuracy and rollout trade-off by explicitly steering rollout selection, rather than relying solely on source-side dataset and training dynamic.

\subsection{Ablation Studies}

In this part, we use Qwen2.5-Math-1.5B trained on the DAPO + MATH dataset to test in detail how L2RR improves the performance and reduces rollout overhead in the training process.

\textbf{Contribution of each component.} We first conduct an ablation study on the contribution of each component in our algorithm in Table~\ref{tab:ablation_components}.
We compare the full L2RR with three ablated variants: \textsc{L2RR w/o ws} (setting $\lambda=0$ to remove the rollout-waste regularizer), \textsc{L2RR w/o rollout allocation} (using a fixed group size $G_{i,t}\equiv G$), and \textsc{L2RR uniform $q$} (fixing the sampler to be uniform and disabling distribution learning).
Overall, the results show that each component plays an important role: removing the rollout-waste term lowers the average accuracy (50.9\% $\rightarrow$ 49.5\%) at a similar rollout count (2.7M vs.\ 3.0M), removing rollout allocation lowers the RO ratio (2.5$\times$ $\rightarrow$ 1.8$\times$) by losing fine-grained budget reallocation within each batch, and fixing $\q$ to uniform leads to the most consistent drop in target accuracy, highlighting the importance of target-guided sampler learning for handling source--target mismatch.
In particular, the full L2RR achieves the best average accuracy while maintaining favorable rollout counts, supporting the contribution of the combined components to target performance and rollout efficiency.

\begin{table*}[t]
\centering
\caption{\textbf{Ablation study on key components of L2RR.}
We report performance (\%) on six math reasoning benchmarks for \textit{Qwen2.5-Math-1.5B} trained on the DAPO + MATH dataset (DM).
\# Rollout denotes cumulative source responses, in millions.
RO ratio is the DAPO count divided by the method count.}
\label{tab:ablation_components}
\setlength{\tabcolsep}{6pt}
\renewcommand{\arraystretch}{1.15}
\resizebox{0.98\textwidth}{!}{
\begin{tabular}{c|cccccc|c|c|c}
\hline
Method & Math500 & AIME24 & AMC & Gaokao & Miner. & Olymp. & Avg. & \# Rollout & RO ratio \\
\hline\hline
L2RR full & \textbf{79.9} & \textbf{18.5} & \textbf{62.2} & \textbf{67.3} & \textbf{35.7} & \textbf{41.5} & \textbf{50.9} & 3.0M & 2.5$\times$ \\
L2RR w/o ws & 78.4 & 16.7 & 61.8 & 66.0 & 34.0 & 40.0 & 49.5 & 2.7M & 2.8$\times$ \\
L2RR w/o rollout allo & 78.2 & 16.7 & 61.5 & 65.7 & 33.8 & 39.8 & 49.3 & 4.3M & 1.8$\times$ \\
L2RR uniform $q$ & 77.9 & 15.7 & 61.8 & 65.2 & 33.0 & 39.5 & 48.9 & 3.2M & 2.4$\times$ \\
\hline
\end{tabular}
}
\vspace{-1mm}
\end{table*}

\begin{table*}[t]
\centering
\caption{\textbf{Effect of held-out target set size for sampler learning.} Accuracy (\%) on six reasoning benchmarks for Qwen2.5-Math-1.5B trained on DAPO + MATH. We vary the size of held-out target dataset to update $\q_t$.}
\label{tab:ablation_targetsize}
\setlength{\tabcolsep}{5pt}
\renewcommand{\arraystretch}{1.12}
\resizebox{0.8\textwidth}{!}{
\begin{tabular}{c|cccccc|c}
\hline
\# Target (held-out) & Math500 & AIME24 & AMC & Gaokao & Miner. & Olymp. & Avg. \\
\hline\hline
4\%  & 78.2 & 16.7 & 61.5 & 65.7 & 34.4 & 39.5 & 49.3 \\
6\%  & 78.5 & 17.6 & 61.8 & 66.0 & 34.7 & 39.8 & 49.7 \\
8\% & 78.9 & 17.6 & 61.5 & 66.3 & 34.9 & 40.1 & 49.9 \\
10\% & 79.9 & 18.5 & 62.2 & 67.3 & 35.7 & 41.5 & 50.9 \\
\hline
\end{tabular}
}
\vspace{-2mm}
\end{table*}

\textbf{Number of held-out target examples.}
We further study how the amount of held-out target data affects L2RR.
Keeping the source training setup fixed, we vary the size of the held-out target dataset used to compute the target-side evaluation objective for updating the sampler $q_t$.
For each size, we uniformly sub-sample the held-out target set, run L2RR with the same training budget, and report the checkpoint with the best average accuracy across the six benchmarks.
Table~\ref{tab:ablation_targetsize} shows that target accuracy increases monotonically with the held-out target size (49.3\% at 4\% to 50.9\% at 10\%); L2RR already benefits from a small held-out slice, and these single runs do not indicate saturation within the range studied.

\begin{figure*}[t]
\centering
\caption{\textbf{Training-dynamics illustration for L2RR.} During training, L2RR increases sampling on prompts that contribute more to improving held-out target accuracy and suppresses repeatedly wasteful prompts as indicated by the history-based counter. Qwen2.5-Math-1.5B trained on DM. DS denotes DAPO with dynamic sampling; DS + ABS adds the adaptive sampling batch size of GRESO~\citep{zheng2025act}.}
\begin{subfigure}[t]{0.32\textwidth}
  \centering
  \includegraphics[width=\linewidth]{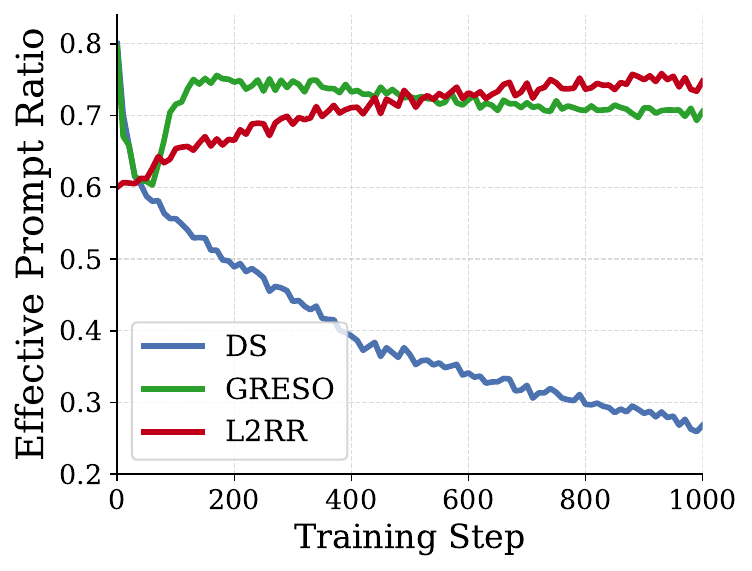}
  \caption{Effective prompt ratio per training step.}
\end{subfigure}\hfill
\begin{subfigure}[t]{0.32\textwidth}
  \centering
  \includegraphics[width=\linewidth]{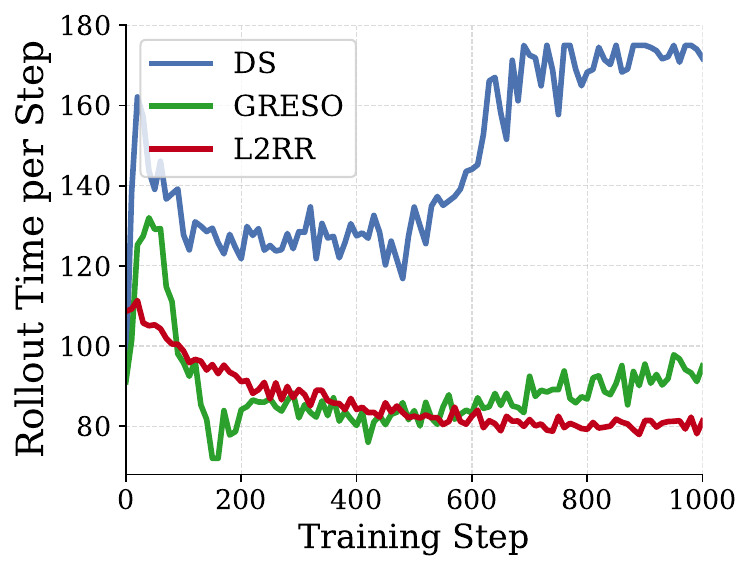}
  \caption{Rollout time per training step (seconds).}
\end{subfigure}\hfill
\begin{subfigure}[t]{0.32\textwidth}
  \centering
  \includegraphics[width=\linewidth]{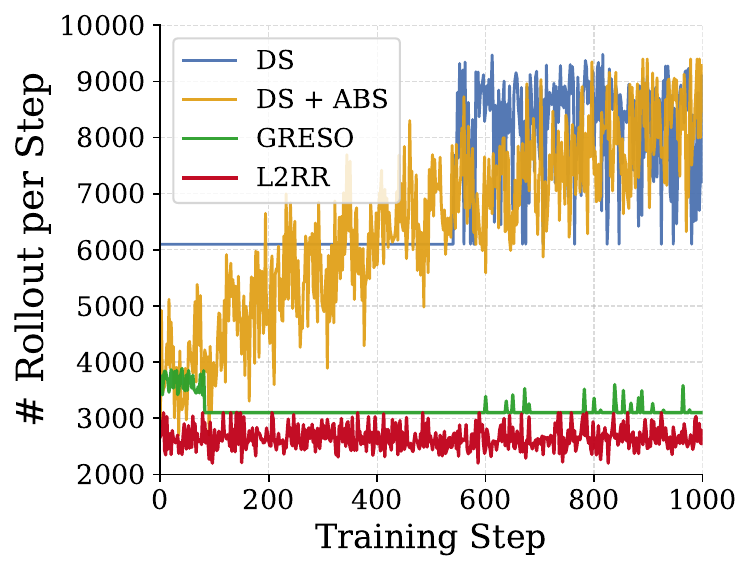}
  \caption{Rollouts per step under dynamic rollout allocation.}
\end{subfigure}
\label{fig:l2rr_dynamics_schematic}
\vspace{-2mm}
\end{figure*}

\textbf{Dynamics of training process.} 
To better understand where the proposed algorithm gains efficiency, we analyze its training dynamics in Figure~\ref{fig:l2rr_dynamics_schematic}.
Overall, L2RR improves both rollout efficiency and target accuracy by coupling target-guided sampler learning with waste-aware rollout allocation.

Figure~\ref{fig:l2rr_dynamics_schematic}(a) reports the effective-prompt ratio over training steps, where an ``effective'' prompt is one whose sampled group does not collapse to a degenerate outcome and thus can contribute non-trivial GRPO updates.
After roughly the first 40 steps, L2RR keeps a much higher effective ratio than DS; GRESO is higher early on, but L2RR overtakes it after about 550 steps and keeps rising, suggesting that the learned sampling distribution increasingly prioritizes prompts that are both target-relevant and less likely to yield repeatedly degenerate behavior, while the rollout-waste regularization discourages spending compute on persistently unproductive prompts.

Figure~\ref{fig:l2rr_dynamics_schematic}(b) shows rollout time per step.
Since L2RR reduces repeated sampling of prompts that historically contribute little training gradients and reallocates effort toward prompts that better improve held-out target accuracy, its rollout time per step stays well below that of DS and, after about 600 steps, is also lower and more stable than that of GRESO, indicating improved utilization of the rollout budget. Accumulating the per-step times in Figure~\ref{fig:l2rr_dynamics_schematic}(b) over 1000 steps, L2RR needs the least total rollout time (about 24 hours, vs.\ about 25 hours for GRESO and 41 hours for DS).

Finally, Figure~\ref{fig:l2rr_dynamics_schematic}(c) visualizes the number of rollouts per step under variable rollout allocation.
Unlike methods that use a fixed group size for every sampled prompt, L2RR dynamically assigns fewer responses to prompts with large waste counters and more responses to prompts that more strongly affect the target-guided sampler update, thereby reducing rollouts per step while maintaining or improving the downstream target accuracy.
This behavior is consistent with our theoretical view that improving the accuracy of the target-driven sampler gradient reduces the effective outer-gradient noise, leading to more stable sampler updates and better target-domain generalization.

\begin{figure*}[t]
\centering
\caption{\textbf{Effect of the rollout-waste weight $\lambda$ on target accuracy (\%).} Overall, moderate $\lambda$ yields the best trade-off, while overly small or large $\lambda$ can under-regularize wasted rollouts or over-penalize informative prompts and thus reduce accuracy.}
\label{fig:ablation_lambda}
\setlength{\tabcolsep}{2pt}
\begin{tabular}{cccc}
\includegraphics[width=0.24\linewidth]{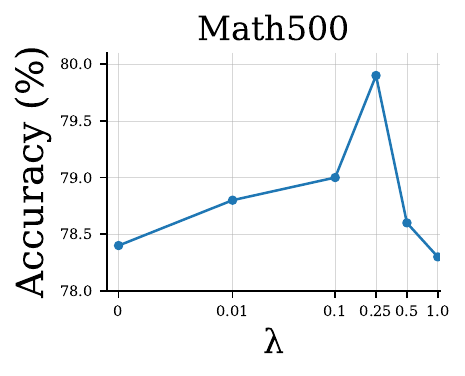} &
\includegraphics[width=0.24\linewidth]{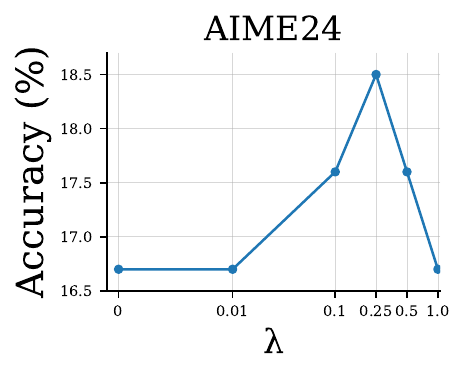} &
\includegraphics[width=0.24\linewidth]{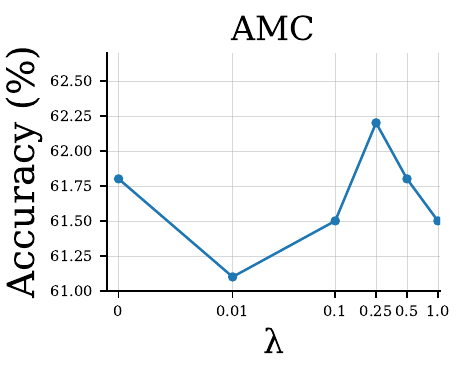} &
\includegraphics[width=0.24\linewidth]{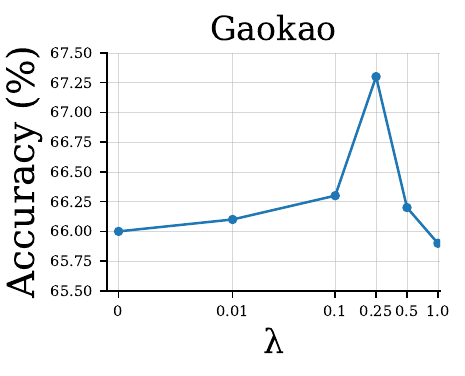}
\end{tabular}
\vspace{-4mm}
\end{figure*}

\textbf{Effect of the rollout-waste weight $\lambda$.}
We also study how the rollout-waste regularizer weight $\lambda$ in Eq.~\eqref{eqn:main_obj} affects target performance.
Intuitively, increasing $\lambda$ strengthens the preference for avoiding prompts that repeatedly induce degenerate GRPO updates, which can improve rollout efficiency but may also reduce exploration that the hard prompt may have positive advantages; setting $\lambda$ too small makes the sampler update rely mainly on the target-guided term and can spend more rollouts on prompts that are unproductive.

Figure~\ref{fig:ablation_lambda} reports accuracy trends on four representative benchmarks as $\lambda$ varies over $\{0, 0.01, 0.1, 0.25, 0.5, 1.0\}$.
Across datasets, we observe a consistent pattern: moderate values of $\lambda$ tend to preserve (or improve) accuracy relative to $\lambda=0$, whereas overly large $\lambda$ can dampen learning on some datasets by aggressively down-weighting a broad set of prompts.
Based on these, we adopt a moderate value of $\lambda$.

\section{Conclusion}
\label{sec:conclusion}
In this paper, we study source-target datasets mismatch in sample-efficient RLVR, where the source training distribution differs from the target evaluation distribution.
We focus on a setting in which the trainer cannot access target prompts or rollout traces, but can obtain aggregated gradient-based feedback from a small held-out target validation split.
We propose \emph{learning-to-reweight rollouts}~{(L2RR)}, which learns a sampling distribution over source prompts to jointly improve target-domain generalization and rollout efficiency in GRPO training.
Specifically, L2RR updates the sampling distribution using exponentiated-gradient steps guided by target-gradient alignment and a rollout-waste regularizer, and uses the resulting sampler to drive GRPO policy updates.
Our theoretical analysis provides a sublinear regret bound for the sampler and a conditional target-loss bound for the policy, supporting the proposed sampling-distribution learning mechanism.
Experiments on six widely-used RLVR benchmarks and three models show that L2RR improves target generalization while increasing rollout efficiency over strong contenders.

\section*{Acknowledgments}
Zhen-Yu Zhang was supported by JSPS KAKENHI Grant Number JP25K21282.
MS was supported by JST ASPIRE Grant Number JPMJAP2405.
This research used computational resources of the supercomputer RIKYU provided by the RIKEN Center for Computational Science.

\bibliography{Refs}

\begin{thebibliography}{10}

\bibitem{bai2024federated}
Jiamu Bai, Daoyuan Chen, Bingchen Qian, Liuyi Yao, and Yaliang Li.
\newblock Federated fine-tuning of large language models under heterogeneous tasks and client resources.
\newblock {\em Advances in Neural Information Processing Systems (NeurIPS)}, 37:14457--14483, 2024.

\bibitem{colson2007overview}
Beno{\^\i}t Colson, Patrice Marcotte, and Gilles Savard.
\newblock An overview of bilevel optimization.
\newblock {\em Annals of operations research}, 153(1):235--256, 2007.

\bibitem{du2024AUF}
Wen-Bo Du, Tian Qin, Tian-Zuo Wang, and Zhi-Hua Zhou.
\newblock Avoiding undesired future with minimal cost in non-stationary environments.
\newblock {\em Advances in Neural Information Processing Systems (NeurIPS)}, 37:135741--135769, 2024.

\bibitem{duchi2011adaptive}
John Duchi, Elad Hazan, and Yoram Singer.
\newblock Adaptive subgradient methods for online learning and stochastic optimization.
\newblock {\em Journal of Machine Learning Research (JMLR)}, 12(7), 2011.

\bibitem{fatemi2025concise}
Mehdi Fatemi, Banafsheh Rafiee, Mingjie Tang, and Kartik Talamadupula.
\newblock Concise reasoning via reinforcement learning.
\newblock {\em arXiv preprint arXiv:2504.05185}, 2025.

\bibitem{fatkhullin2024taming}
Ilyas Fatkhullin and Niao He.
\newblock Taming nonconvex stochastic mirror descent with general {Bregman} divergence.
\newblock In {\em Proceedings of the 27th International Conference on Artificial Intelligence and Statistics (AISTATS)}, pages 3493--3501, 2024.

\bibitem{franceschi2018bilevel}
Luca Franceschi, Paolo Frasconi, Saverio Salzo, Riccardo Grazzi, and Massimiliano Pontil.
\newblock Bilevel programming for hyperparameter optimization and meta-learning.
\newblock In {\em Proceedings of the 35th International Conference on Machine Learning (ICML)}, pages 1568--1577, 2018.

\bibitem{guo2025deepseek}
Daya Guo, Dejian Yang, Haowei Zhang, Junxiao Song, Ruoyu Zhang, Runxin Xu, Qihao Zhu, Shirong Ma, Peiyi Wang, Xiao Bi, et~al.
\newblock Deepseek-r1: Incentivizing reasoning capability in llms via reinforcement learning.
\newblock {\em arXiv preprint arXiv:2501.12948}, 2025.

\bibitem{guyon2019analysis}
Isabelle Guyon, Lisheng Sun-Hosoya, Marc Boull{\'e}, Hugo~Jair Escalante, Sergio Escalera, Zhengying Liu, Damir Jajetic, Bisakha Ray, Mehreen Saeed, Mich{\`e}le Sebag, et~al.
\newblock Analysis of the {AutoML} challenge series.
\newblock {\em Automated Machine Learning}, 177:177--219, 2019.

\bibitem{hazan2022online}
Elad Hazan.
\newblock {\em Introduction to Online Convex Optimization}.
\newblock The MIT Press, 2 edition, 2022.

\bibitem{he2024olympiadbench}
Chaoqun He, Renjie Luo, Yuzhuo Bai, Shengding Hu, Zhen Thai, Junhao Shen, Jinyi Hu, Xu~Han, Yujie Huang, Yuxiang Zhang, et~al.
\newblock Olympiadbench: A challenging benchmark for promoting agi with olympiad-level bilingual multimodal scientific problems.
\newblock In {\em Proceedings of the 62nd Annual Meeting of the Association for Computational Linguistics (ACL)}, pages 3828--3850, 2024.

\bibitem{hendrycks2measuring}
Dan Hendrycks, Collin Burns, Saurav Kadavath, Akul Arora, Steven Basart, Eric Tang, Dawn Song, and Jacob Steinhardt.
\newblock Measuring mathematical problem solving with the math dataset.
\newblock In {\em Thirty-fifth Conference on Neural Information Processing Systems Datasets and Benchmarks Track (Round 2)}, 2021.

\bibitem{ivison2025large}
Hamish Ivison, Muru Zhang, Faeze Brahman, Pang~Wei Koh, and Pradeep Dasigi.
\newblock Large-scale data selection for instruction tuning.
\newblock {\em arXiv preprint arXiv:2503.01807}, 2025.

\bibitem{jaech2024openai}
Aaron Jaech, Adam Kalai, Adam Lerer, Adam Richardson, Ahmed El-Kishky, Aiden Low, Alec Helyar, Aleksander Madry, Alex Beutel, Alex Carney, et~al.
\newblock Openai o1 system card.
\newblock {\em arXiv preprint arXiv:2412.16720}, 2024.

\bibitem{khanduri2021near}
Prashant Khanduri, Siliang Zeng, Mingyi Hong, Hoi-To Wai, Zhaoran Wang, and Zhuoran Yang.
\newblock A near-optimal algorithm for stochastic bilevel optimization via double-momentum.
\newblock {\em Advances in Neural Information Processing Systems (NeurIPS)}, 34:30271--30283, 2021.

\bibitem{kwon2023efficient}
Woosuk Kwon, Zhuohan Li, Siyuan Zhuang, Ying Sheng, Lianmin Zheng, Cody~Hao Yu, Joseph Gonzalez, Hao Zhang, and Ion Stoica.
\newblock Efficient memory management for large language model serving with pagedattention.
\newblock In {\em Proceedings of the 29th symposium on operating systems principles}, pages 611--626, 2023.

\bibitem{lambert2024tulu}
Nathan Lambert, Jacob Morrison, Valentina Pyatkin, Shengyi Huang, Hamish Ivison, Faeze Brahman, Lester James~V Miranda, Alisa Liu, Nouha Dziri, Shane Lyu, et~al.
\newblock Tulu 3: Pushing frontiers in open language model post-training.
\newblock {\em arXiv preprint arXiv:2411.15124}, 2024.

\bibitem{lewkowycz2022solving}
Aitor Lewkowycz, Anders Andreassen, David Dohan, Ethan Dyer, Henryk Michalewski, Vinay Ramasesh, Ambrose Slone, Cem Anil, Imanol Schlag, Theo Gutman-Solo, et~al.
\newblock Solving quantitative reasoning problems with language models.
\newblock {\em Advances in Neural Information Processing Systems (NeurIPS)}, 35:3843--3857, 2022.

\bibitem{li2026learning}
Shuozhe Li, Vaishnav Tadiparthi, Kwonjoon Lee, Nakul Agarwal, Hossein~Nourkhiz Mahjoub, Ehsan~Moradi Pari, Lizhang Chen, Amy Zhang, and Liu Leqi.
\newblock Learning robust reasoning through guided adversarial self-play.
\newblock {\em arXiv preprint arXiv:2602.00173}, 2026.

\bibitem{li2025limr}
Xuefeng Li, Haoyang Zou, and Pengfei Liu.
\newblock Limr: Less is more for rl scaling.
\newblock {\em arXiv preprint arXiv:2502.11886}, 2025.

\bibitem{li2022alphacode}
Yujia Li, David Choi, Junyoung Chung, Nate Kushman, Julian Schrittwieser, R{\'e}mi Leblond, Tom Eccles, James Keeling, Felix Gimeno, Agustin Dal~Lago, Thomas Hubert, Peter Choy, Cyprien de~Masson~d'Autume, Igor Babuschkin, Xinyun Chen, Po-Sen Huang, Johannes Welbl, Sven Gowal, Alexey Cherepanov, James Molloy, Daniel~J. Mankowitz, Esme Sutherland~Robson, Pushmeet Kohli, Nando de~Freitas, Koray Kavukcuoglu, and Oriol Vinyals.
\newblock Competition-level code generation with {AlphaCode}.
\newblock {\em Science}, 2022.

\bibitem{liu2023evalplus}
Jiawei Liu, Chunqiu~Steven Xia, Yuyao Wang, and Lingming Zhang.
\newblock Is your code generated by {ChatGPT} really correct? rigorous evaluation of large language models for code generation.
\newblock {\em Advances in Neural Information Processing Systems (NeurIPS)}, 36:21558--21572, 2023.

\bibitem{liu2025understanding}
Zichen Liu, Changyu Chen, Wenjun Li, Penghui Qi, Tianyu Pang, Chao Du, Wee~Sun Lee, and Min Lin.
\newblock Understanding r1-zero-like training: A critical perspective.
\newblock {\em arXiv preprint arXiv:2503.20783}, 2025.

\bibitem{maclaurin2015gradient}
Dougal Maclaurin, David Duvenaud, and Ryan Adams.
\newblock Gradient-based hyperparameter optimization through reversible learning.
\newblock In {\em Proceedings of the 32nd International Conference on Machine Learning (ICML)}, pages 2113--2122, 2015.

\bibitem{muennighoff2025s1}
Niklas Muennighoff, Zitong Yang, Weijia Shi, Xiang~Lisa Li, Li~Fei-Fei, Hannaneh Hajishirzi, Luke Zettlemoyer, Percy Liang, Emmanuel Cand{\`e}s, and Tatsunori~B Hashimoto.
\newblock s1: Simple test-time scaling.
\newblock In {\em Proceedings of the 2025 Conference on Empirical Methods in Natural Language Processing (EMNLP)}, pages 20286--20332, 2025.

\bibitem{muldrew2024active}
William Muldrew, Peter Hayes, Mingtian Zhang, and David Barber.
\newblock Active preference learning for large language models.
\newblock In {\em Proceedings of the 41st International Conference on Machine Learning (ICML)}, pages 36577--36590, 2024.

\bibitem{nemirovski2009robust}
Arkadi Nemirovski, Anatoli Juditsky, Guanghui Lan, and Alexander Shapiro.
\newblock Robust stochastic approximation approach to stochastic programming.
\newblock {\em SIAM Journal on Optimization}, 19(4):1574--1609, 2009.

\bibitem{nguyen2025lightspeed}
Khai Nguyen, Hai Nguyen, Tuan Pham, and Nhat Ho.
\newblock Lightspeed geometric dataset distance via sliced optimal transport.
\newblock In {\em Proceedings of the 42nd International Conference on Machine Learning (ICML)}, pages 46162--46177, 2025.

\bibitem{ouyang2022training}
Long Ouyang, Jeffrey Wu, Xu~Jiang, Diogo Almeida, Carroll Wainwright, Pamela Mishkin, Chong Zhang, Sandhini Agarwal, Katarina Slama, Alex Ray, et~al.
\newblock Training language models to follow instructions with human feedback.
\newblock {\em Advances in Neural Information Processing Systems (NeurIPS)}, 35:27730--27744, 2022.

\bibitem{pillutla2021mauve}
Krishna Pillutla, Swabha Swayamdipta, Rowan Zellers, John Thickstun, Sean Welleck, Yejin Choi, and Zaid Harchaoui.
\newblock {MAUVE}: Measuring the gap between neural text and human text using divergence frontiers.
\newblock {\em Advances in Neural Information Processing Systems (NeurIPS)}, 34:4816--4828, 2021.

\bibitem{rein2024gpqa}
David Rein, Betty~Li Hou, Asa~Cooper Stickland, Jackson Petty, Richard~Yuanzhe Pang, Julien Dirani, Julian Michael, and Samuel~R Bowman.
\newblock Gpqa: A graduate-level google-proof q\&a benchmark.
\newblock In {\em Proceedings of the 1st Conference on Language Modeling}, 2024.

\bibitem{ren2018learning}
Mengye Ren, Wenyuan Zeng, Bin Yang, and Raquel Urtasun.
\newblock Learning to reweight examples for robust deep learning.
\newblock In {\em Proceedings of the 35th International Conference on Machine Learning (ICML)}, pages 4334--4343, 2018.

\bibitem{shao2024deepseekmath}
Zhihong Shao, Peiyi Wang, Qihao Zhu, Runxin Xu, Junxiao Song, Xiao Bi, Haowei Zhang, Mingchuan Zhang, YK~Li, Yang Wu, et~al.
\newblock Deepseekmath: Pushing the limits of mathematical reasoning in open language models.
\newblock {\em arXiv preprint arXiv:2402.03300}, 2024.

\bibitem{sheng2025hybridflow}
Guangming Sheng, Chi Zhang, Zilingfeng Ye, Xibin Wu, Wang Zhang, Ru~Zhang, Yanghua Peng, Haibin Lin, and Chuan Wu.
\newblock Hybridflow: A flexible and efficient rlhf framework.
\newblock In {\em Proceedings of the 20th European Conference on Computer Systems}, pages 1279--1297, 2025.

\bibitem{shu2019meta}
Jun Shu, Qi~Xie, Lixuan Yi, Qian Zhao, Sanping Zhou, Zongben Xu, and Deyu Meng.
\newblock Meta-weight-net: Learning an explicit mapping for sample weighting.
\newblock {\em Advances in Neural Information Processing Systems (NeurIPS)}, 32:1917--1928, 2019.

\bibitem{sugiyama2007covariate}
Masashi Sugiyama, Matthias Krauledat, and Klaus-Robert M{\"u}ller.
\newblock Covariate shift adaptation by importance weighted cross validation.
\newblock {\em Journal of Machine Learning Research (JMLR)}, 8(5):985--1005, 2007.

\bibitem{wang2024embedding}
Yiming Wang, Pei Zhang, Baosong Yang, Derek Wong, Zhuosheng Zhang, and Rui Wang.
\newblock Embedding trajectory for out-of-distribution detection in mathematical reasoning.
\newblock {\em Advances in Neural Information Processing Systems (NeurIPS)}, 37:42965--42999, 2024.

\bibitem{wang2025reinforcement}
Yiping Wang, Qing Yang, Zhiyuan Zeng, Liliang Ren, Liyuan Liu, Baolin Peng, Hao Cheng, Xuehai He, Kuan Wang, Jianfeng Gao, et~al.
\newblock Reinforcement learning for reasoning in large language models with one training example.
\newblock {\em arXiv preprint arXiv:2504.20571}, 2025.

\bibitem{xia2024less}
Mengzhou Xia, Sadhika Malladi, Suchin Gururangan, Sanjeev Arora, and Danqi Chen.
\newblock Less: selecting influential data for targeted instruction tuning.
\newblock In {\em Proceedings of the 41st International Conference on Machine Learning (ICML)}, pages 54104--54132, 2024.

\bibitem{xiong2025minimalist}
Wei Xiong, Jiarui Yao, Yuhui Xu, Bo~Pang, Lei Wang, Doyen Sahoo, Junnan Li, Nan Jiang, Tong Zhang, Caiming Xiong, et~al.
\newblock A minimalist approach to llm reasoning: from rejection sampling to reinforce.
\newblock {\em arXiv preprint arXiv:2504.11343}, 2025.

\bibitem{yang2024qwen2}
An~Yang, Beichen Zhang, Binyuan Hui, Bofei Gao, Bowen Yu, Chengpeng Li, Dayiheng Liu, Jianhong Tu, Jingren Zhou, Junyang Lin, et~al.
\newblock Qwen2. 5-math technical report: Toward mathematical expert model via self-improvement.
\newblock {\em arXiv preprint arXiv:2409.12122}, 2024.

\bibitem{ye2024fedllm}
Rui Ye, Rui Ge, Xinyu Zhu, Jingyi Chai, Du~Yaxin, Yang Liu, Yanfeng Wang, and Siheng Chen.
\newblock Fedllm-bench: Realistic benchmarks for federated learning of large language models.
\newblock {\em Advances in Neural Information Processing Systems (NeurIPS)}, 37:111106--111130, 2024.

\bibitem{yu2025dapo}
Qiying Yu, Zheng Zhang, Ruofei Zhu, Yufeng Yuan, Xiaochen Zuo, Yu~Yue, Weinan Dai, Tiantian Fan, Gaohong Liu, Lingjun Liu, et~al.
\newblock Dapo: An open-source llm reinforcement learning system at scale.
\newblock {\em arXiv preprint arXiv:2503.14476}, 2025.

\bibitem{yuan2025s}
Yufeng Yuan, Yu~Yue, Ruofei Zhu, Tiantian Fan, and Lin Yan.
\newblock What's behind ppo's collapse in long-cot? value optimization holds the secret.
\newblock {\em arXiv preprint arXiv:2503.01491}, 2025.

\bibitem{zhang2023evaluating}
Xiaotian Zhang, Chunyang Li, Yi~Zong, Zhengyu Ying, Liang He, and Xipeng Qiu.
\newblock Evaluating the performance of large language models on gaokao benchmark.
\newblock {\em arXiv preprint arXiv:2305.12474}, 2023.

\bibitem{zhang2025improving}
Yuheng Zhang, Wenlin Yao, Changlong Yu, Yao Liu, Qingyu Yin, Bing Yin, Hyokun Yun, and Lihong Li.
\newblock Improving sampling efficiency in rlvr through adaptive rollout and response reuse.
\newblock {\em arXiv preprint arXiv:2509.25808}, 2025.

\bibitem{zhang2026consolidating}
Zheyu Zhang, Shuo Yang, and Gjergji Kasneci.
\newblock Consolidating rewarded perturbations for llm post-training.
\newblock {\em arXiv preprint arXiv:2605.31494}, 2026.

\bibitem{zheng2025act}
Haizhong Zheng, Yang Zhou, Brian~R Bartoldson, Bhavya Kailkhura, Fan Lai, Jiawei Zhao, and Beidi Chen.
\newblock Act only when it pays: Efficient reinforcement learning for llm reasoning via selective rollouts.
\newblock {\em Advances in Neural Information Processing Systems (NeurIPS)}, 38:in press, 2025.

\end{thebibliography}
\bibliographystyle{plain}

\newpage

\appendix

\onecolumn


\newpage

\onecolumn

\newcommand{\GRPO}{\mathrm{GRPO}}
\newcommand{\adv}{\mathrm{ws}}
\newcommand{\KL}{\mathrm{KL}}
\newcommand{\DeltaS}{\Delta^{N_S}}

\section*{Appendix}

\section{Detailed Experimental Setting}
\label{app:exp_details}
Table~\ref{tab:hyperparams} lists the hyperparameters.
Shared GRPO settings follow~\citep{zheng2025act} and are identical for all methods; contenders use their published method-specific settings.

\begin{table}[h]
\centering
\caption{\textbf{Hyperparameters.} Batch sizes count prompts.}
\label{tab:hyperparams}
\renewcommand{\arraystretch}{1.1}
\small
\begin{tabular}{ll}
\hline
\multicolumn{2}{c}{\normalsize Training} \\
Framework & verl (FSDP) with vLLM rollouts \\
Optimizer & AdamW, $\beta=(0.9,0.999)$, weight decay $0.01$ \\
Learning rate $\eta_\theta$ & $1\times10^{-6}$, constant \\
Training steps & 1000 \\
Batch size $B$ & 256 (DeepSeek-1.5B: 128) \\
Prompts per sampling round & 384 (DeepSeek-1.5B: 192) \\
PPO mini-batch size & 64 (4 updates per step; DeepSeek-1.5B: 2) \\
Group size $G$ & 8 \\
Loss & token-level GRPO, $\epsilon_{\text{low}}=\epsilon_{\text{high}}=0.2$, entropy coefficient $0.001$, no KL \\
Max prompt / response length & Qwen2.5-Math: 1536 / 2560; DeepSeek-1.5B: 2048 / 6144 \\
Rollout temperature & 1.0 \\
Reward & binary: 1 if correct, 0 otherwise (Math-Verify) \\
\hline
\multicolumn{2}{c}{\normalsize Evaluation} \\
Frequency & every 50 steps on the validation splits \\
Decoding & temperature 1.0, top-$p$ 0.7, avg@4 \\
Checkpoint selection & best macro-average validation accuracy \\
Validation / test split & 10\% / 90\% of each benchmark \\
\hline
\multicolumn{2}{c}{\normalsize L2RR} \\
Rollout-waste weight $\lambda$ & 0.25 \\
Sampler step size $\eta_q$ & 1.0 \\
Waste decay $p_e$ & 0.5 \\
Minimum group size $G_{\min}$ & 4 \\
Initial sampler $\q_1$ & uniform \\
Sampler floor & mixing with uniform, floor 0.02; cap $10\times$ uniform \\
Target-gradient batch & 64 validation prompts $\times$ 8 rollouts, every step \\
Target-term normalization & RMS, clipped at 2.0 \\
\hline
\multicolumn{2}{c}{\normalsize Compute} \\
Hardware & 2 NVIDIA GB200 GPUs \\
\hline
\end{tabular}
\end{table}

\section{More Empirical Results}
\label{app:emprical_fair}
In this section, we provide detailed empirical results omitted from the main paper due to space limits.

\subsection{Comparison with Target-Aware Baselines.} 
To further ensure a fair comparison, we allow DAPO, GRESO, and AR3PO to use the same held-out target data as L2RR.
For each contender, we evaluate checkpoints every 50 steps during training and use the held-out target set for target-validation-based checkpoint selection.
As shown in Table~\ref{tab:perf_rollouts_all}, L2RR consistently achieves the best average performance across different models and source datasets.
Compared with the original source-only contenders, both Target-Mix and TargetVal variants generally improve target-domain performance, confirming that held-out target information is useful under source-target mismatch.
However, the performance of Target-Mix and TargetVal is largely comparable, and both remain inferior to L2RR.
This suggests that the gains of L2RR do not simply come from access to additional target data; rather, explicitly using target-side feedback to learn the source-prompt sampling distribution provides a more effective way to exploit limited target information than directly mixing target examples into training or using them only for model selection.

\subsection{Comparison with Direct Target Training}
\label{app:target_only}

To examine how target information is used, we compare L2RR with \textsc{Target-only}, which performs GRPO policy training exclusively on the target validation split, and \textsc{Target-Mix} variants in Table~\ref{tab:perf_rollouts_all}, which mix this split with source data for training.
Both baselines require direct access to target prompts and therefore exceed the target-feedback-only access assumed by L2RR.

Table~\ref{tab:target_only} shows that training on the small target split alone yields substantially lower accuracy, while L2RR also outperforms the best Target-Mix variant in each configuration.
These comparisons examine different uses of target information.

\begin{table}[h]
\centering
\caption{\textbf{Diagnostic comparison with direct target training.} We report average test accuracy (\%) across the six math reasoning benchmarks. Best Target-Mix is the highest average accuracy among the DAPO, GRESO, and AR3PO Target-Mix variants in Table~\ref{tab:perf_rollouts_all}. Target-only uses no source data, so its result is repeated across the DM and OR1 rows for each model.}
\label{tab:target_only}
\setlength{\tabcolsep}{7pt}
\renewcommand{\arraystretch}{1.1}
\resizebox{0.8\textwidth}{!}{
\begin{tabular}{l|c|ccc}
\hline
Model & Source & Target-only & Best Target-Mix & L2RR \\
\hline\hline
\multirow{2}{*}{Qwen2.5-Math-1.5B}
 & DM & 39.3 & 50.4 & \textbf{50.9} \\
 & OR1 & 39.3 & 47.7 & \textbf{48.7} \\
\hline
\multirow{2}{*}{DeepSeek-R1-Distill-Qwen-1.5B}
 & DM & 50.1 & 61.6 & \textbf{62.4} \\
 & OR1 & 50.1 & 58.1 & \textbf{59.1} \\
\hline
\multirow{2}{*}{Qwen2.5-Math-7B}
 & DM & 47.4 & 58.7 & \textbf{59.3} \\
 & OR1 & 47.4 & 56.0 & \textbf{56.8} \\
\hline
\end{tabular}
}
\vspace{-2mm}
\end{table}

\subsection{Comparison with Simple Target-Guided Baselines}
\label{app:simple_target_guided}

We compare L2RR with two simple target-guided baselines using Qwen2.5-Math-1.5B trained on DM and OR1.
Both baselines sample source prompts uniformly and use the same target-feedback interface and validation-based checkpoint-selection protocol as L2RR, with evaluation on the disjoint test splits.

\textbf{Gradient-Similarity Thresholding.} For each source prompt in the current batch, we compute the alignment score $s_{i,t}=\langle g^{\rmtgt}_{t+1},g_{t,i}-\bar g_t\rangle$, where $\bar g_t=B^{-1}\sum_{j\in\B_t}g_{t,j}$ and the target gradient is evaluated at the same look-ahead checkpoint used by L2RR.
Only prompts with positive scores are retained in the policy update, with equal weights.

\textbf{Target-Loss Reweighting.} Following the one-step reweighting strategy of~\citep{ren2018learning}, we form a virtual update $\theta'_t(\epsilon)=\theta_t-\eta_\theta\sum_{j\in\B_t}\epsilon_jg_{t,j}$ and assign weights proportional to $[-\partial L_{\rmtgt}(\theta'_t(\epsilon))/\partial\epsilon_i\vert_{\epsilon=0}]_+$.
We normalize the positive weights to sum to one and use them in the policy update, falling back to uniform weights when all weights are zero.
Both baselines operate only on the current batch, without learning a persistent sampler or adaptively allocating rollouts.

\textbf{Rollout budget.} We keep the training steps and source batch size fixed and choose a fixed per-prompt group size for each baseline to match L2RR's cumulative source-rollout count.
Group sizes are selected from the same feasible range based only on rollout counts.
Rollouts discarded by Gradient-Similarity Thresholding are included in the count.

As shown in Table~\ref{tab:simple_target_guided}, L2RR improves over Gradient-Similarity Thresholding 1.0 percentage point on both sources and over Target-Loss Reweighting by 0.9/0.8 points on DM/OR1, respectively.
These results support the combined design of persistent sampler learning, history-dependent rollout-waste control, and adaptive rollout allocation beyond using target feedback for within-batch filtering or reweighting alone.

\begin{table}[h]
\centering
\caption{\textbf{Comparison with simple target-guided baselines.} We report average test accuracy (\%) across the six math reasoning benchmarks and cumulative source-rollout counts for Qwen2.5-Math-1.5B. All methods use the same target-feedback interface.}
\label{tab:simple_target_guided}
\setlength{\tabcolsep}{8pt}
\renewcommand{\arraystretch}{1.1}
\resizebox{0.8\textwidth}{!}{
\begin{tabular}{c|l|c|c}
\hline
Source & Method & Avg. (\%) & Total rollouts \\
\hline\hline
\multirow{3}{*}{DM}
 & Gradient-Similarity Thresholding & 49.9 & 3.0M \\
 & Target-Loss Reweighting & 50.0 & 3.0M \\
 & L2RR & \textbf{50.9} & 3.0M \\
\hline
\multirow{3}{*}{OR1}
 & Gradient-Similarity Thresholding & 47.7 & 1.4M \\
 & Target-Loss Reweighting & 47.9 & 1.4M \\
 & L2RR & \textbf{48.7} & 1.4M \\
\hline
\end{tabular}
}
\vspace{-2mm}
\end{table}

\subsection{Quantifying Source--Target Mismatch}
\label{app:source_target_mismatch}

We quantify source--target mismatch for the 12 pairs formed by the two source datasets and six target benchmarks using s-OTDD~\citep{nguyen2025lightspeed} and $1-\mathrm{MAUVE}$~\citep{pillutla2021mauve}.
Both are oriented so that larger values indicate greater mismatch.
We report $1-\mathrm{MAUVE}$ because MAUVE itself measures distributional similarity.

For each pair, we sample $\min(512,N)$ prompts from each side, where $N$ is the number of prompts in the target benchmark, and average the distance estimates over five independent source subsamples.
Both metrics use the same prompt representations from the frozen sentence encoder \texttt{all-mpnet-base-v2}, which is not updated during training.
For s-OTDD, we assign difficulty labels using each prompt's empirical success rate over 16 responses sampled from a frozen pretrained model.
We compute difficulty tertiles separately within each dataset to obtain easy, medium, and hard groups.
The $1-\mathrm{MAUVE}$ metric uses the same representations without difficulty labels.
This analysis is entirely post hoc, and neither the distance estimates nor the difficulty labels are used for policy training, sampler updates, checkpoint selection, or hyperparameter tuning.

For each model and source--target pair, we define the gain $\Delta$ as L2RR's mean test accuracy minus the highest mean test accuracy among DAPO, GRESO, and AR3PO, using the five-run means in Table~\ref{tab:perf_rollouts}.
Table~\ref{tab:source_target_mismatch} reports the distances and gains in percentage points.

\begin{table}[h]
\centering
\caption{\textbf{Source--target mismatch and L2RR gains.} Distances are averaged over five source subsamples. Gains are computed from the five-run mean test accuracies in Table~\ref{tab:perf_rollouts}. Qwen-1.5B, DeepSeek-1.5B, and Qwen-7B denote Qwen2.5-Math-1.5B, DeepSeek-R1-Distill-Qwen-1.5B, and Qwen2.5-Math-7B, respectively.}
\label{tab:source_target_mismatch}
\setlength{\tabcolsep}{6pt}
\renewcommand{\arraystretch}{1.1}
\resizebox{0.9\textwidth}{!}{
\begin{tabular}{c|l|cc|ccc}
\hline
Source & Target & s-OTDD & $1-\mathrm{MAUVE}$ & $\Delta$ Qwen-1.5B & $\Delta$ DeepSeek-1.5B & $\Delta$ Qwen-7B \\
\hline\hline
\multirow{6}{*}{DM}
 & Math500 & 0.235 & 0.194 & $+1.8$ & $+1.8$ & $+1.2$ \\
 & AIME24 & 0.369 & 0.303 & $+1.5$ & $+1.9$ & $-0.9$ \\
 & AMC & 0.204 & 0.175 & $+0.9$ & $+1.3$ & $+0.9$ \\
 & Gaokao & 0.290 & 0.238 & $+1.3$ & $+1.4$ & $+0.5$ \\
 & Minerva & 0.302 & 0.288 & $-0.2$ & $+0.9$ & $+1.2$ \\
 & OlympiadBench & 0.422 & 0.379 & $+2.8$ & $+2.5$ & $+2.1$ \\
\hline
\multirow{6}{*}{OR1}
 & Math500 & 0.298 & 0.248 & $+1.8$ & $+1.1$ & $+0.9$ \\
 & AIME24 & 0.358 & 0.320 & $+1.3$ & $+1.8$ & $+1.5$ \\
 & AMC & 0.206 & 0.162 & $+1.3$ & $-0.4$ & $+1.3$ \\
 & Gaokao & 0.274 & 0.240 & $+1.1$ & $+1.7$ & $+1.2$ \\
 & Minerva & 0.311 & 0.286 & $+0.4$ & $+1.4$ & $+0.7$ \\
 & OlympiadBench & 0.422 & 0.382 & $+2.3$ & $+3.1$ & $+1.4$ \\
\hline
\end{tabular}
}
\vspace{-2mm}
\end{table}

The relationship between mismatch and gain is model-dependent.
Across the 12 pairs, the Pearson correlations with s-OTDD/$1-\mathrm{MAUVE}$ are 0.48/0.37, 0.76/0.75, and 0.08/0.17 for Qwen-1.5B, DeepSeek-1.5B, and Qwen-7B, respectively.
The association is strongest for DeepSeek-1.5B and weak for Qwen-7B.
These descriptive correlations concern a small set of pairs sharing source datasets and target benchmarks, and should not be interpreted as a causal relationship or a guarantee of gains on new datasets.

\subsection{Sensitivity to the Rollout-Waste Weight}
\label{app:lambda_sensitivity}

We extend the rollout-waste weight ablation with a finer sweep over $\lambda\in\{0.10,0.15,0.20,0.25,0.30,0.40,0.50\}$, using Qwen2.5-Math-1.5B trained on DM.
Table~\ref{tab:lambda_fine_grid} reports accuracy on four target benchmarks and includes $\lambda=0.01$ and $1.00$ for comparison.

\begin{table}[h]
\centering
\caption{\textbf{Fine-grained sensitivity analysis of the rollout-waste weight $\lambda$.} We report accuracy (\%) for Qwen2.5-Math-1.5B trained on DM. Bold indicates the highest reported accuracy for each benchmark, including ties.}
\label{tab:lambda_fine_grid}
\setlength{\tabcolsep}{6pt}
\renewcommand{\arraystretch}{1.1}
\resizebox{0.8\textwidth}{!}{
\begin{tabular}{l|ccccccccc}
\hline
Benchmark / $\lambda$ & 0.01 & 0.10 & 0.15 & 0.20 & 0.25 & 0.30 & 0.40 & 0.50 & 1.00 \\
\hline\hline
Math500 & 78.8 & 79.0 & 79.1 & 79.3 & \textbf{79.9} & 79.5 & 78.7 & 78.6 & 78.3 \\
AIME24 & 16.7 & 17.6 & 17.6 & \textbf{18.5} & \textbf{18.5} & \textbf{18.5} & 17.6 & 17.6 & 16.7 \\
AMC & 61.1 & 61.5 & 61.5 & 61.8 & \textbf{62.2} & \textbf{62.2} & 61.8 & 61.8 & 61.5 \\
Gaokao & 66.1 & 66.3 & 66.3 & 66.3 & \textbf{67.3} & 66.5 & 66.1 & 66.2 & 65.9 \\
\hline
\end{tabular}
}
\vspace{-2mm}
\end{table}

All four benchmarks peak near $\lambda=0.25$, and moderate values in $[0.10,0.50]$ stay within 1.3, 0.9, 0.7, and 1.2 percentage points of the peak on Math500, AIME24, AMC, and Gaokao, respectively, while the lowest values occur at $\lambda=0.01$ or $\lambda=1.00$. Each $\lambda$ value is a single run, and all of these gaps are smaller than the five-seed standard deviations of L2RR in this setting (1.7--2.4 points; Table~\ref{tab:perf_rollouts}), so the sweep supports a broad moderate range around $\lambda=0.25$ rather than a sharply tuned optimum.

\subsection{Sensitivity to Target Validation Bias}
\label{app:validation_bias}

We examine how biased target validation data affect L2RR when trained on DM and OR1.
We consider difficulty bias on Math500 and Gaokao and subdomain bias on OlympiadBench.
In each setting, the target validation set contains 10\% of the benchmark examples, and the test examples are disjoint from the validation set.

\textbf{Difficulty bias.} For Math500, we group difficulty levels 1--2 as easy, level 3 as medium, and levels 4--5 as hard.
For Gaokao, we use the official per-question score as a prespecified difficulty proxy, with higher-scoring questions treated as harder.
Uniform validation samples examples uniformly from the benchmark.
Easy-biased and Hard-biased validation use easy/medium/hard proportions of 70\%/20\%/10\% and 10\%/20\%/70\%, respectively.
For each condition, we sample the test set from the remaining examples according to the original benchmark's empirical difficulty proportions.

\begin{table}[h]
\centering
\caption{\textbf{Sensitivity to difficulty-biased target validation data.} Accuracy is reported in percent. $\Delta$ is the change in the average accuracy over Math500 and Gaokao relative to Uniform for the same source dataset, in percentage points.}
\label{tab:validation_difficulty_bias}
\setlength{\tabcolsep}{8pt}
\renewcommand{\arraystretch}{1.1}
\resizebox{0.8\textwidth}{!}{
\begin{tabular}{c|l|cc|c|c}
\hline
Source & Validation set & Math500 & Gaokao & $\Delta$ & Total rollouts \\
\hline\hline
\multirow{3}{*}{DM}
 & Uniform & 79.9 & 67.3 & -- & 3.0M \\
 & Easy-biased & 80.0 & 67.6 & $+0.2$ & 2.8M \\
 & Hard-biased & 79.8 & 66.7 & $-0.4$ & 3.3M \\
\hline
\multirow{3}{*}{OR1}
 & Uniform & 78.8 & 66.7 & -- & 1.4M \\
 & Easy-biased & 78.9 & 65.8 & $-0.4$ & 1.3M \\
 & Hard-biased & 78.1 & 66.6 & $-0.4$ & 1.6M \\
\hline
\end{tabular}
}
\vspace{-2mm}
\end{table}

\textbf{Subdomain bias.}
We use OlympiadBench's Algebra, Geometry, Number Theory, and Combinatorics labels.
Each biased validation set contains 70\% examples from the designated subdomain and 30\% from the other subdomains combined.
For each condition, we sample the test set from the remaining examples according to the original benchmark's empirical subdomain proportions.
Thus, these bias experiments preserve the original target group proportions in the test sampling, but do not use an identical fixed test set across conditions.

\begin{table}[h]
\centering
\caption{\textbf{Sensitivity to subdomain-biased target validation data.} We report OlympiadBench accuracy (\%), its change $\Delta$ relative to Uniform for the same source dataset in percentage points, and cumulative rollout counts.}
\label{tab:validation_subdomain_bias}
\setlength{\tabcolsep}{8pt}
\renewcommand{\arraystretch}{1.1}
\resizebox{0.8\textwidth}{!}{
\begin{tabular}{c|l|c|c|c}
\hline
Source & Validation set & OlympiadBench & $\Delta$ & Total rollouts \\
\hline\hline
\multirow{5}{*}{DM}
 & Uniform & 41.5 & -- & 3.0M \\
 & Algebra-biased & 41.0 & $-0.5$ & 3.1M \\
 & Geometry-biased & 41.8 & $+0.3$ & 2.9M \\
 & Number-Theory-biased & 41.2 & $-0.3$ & 3.1M \\
 & Combinatorics-biased & 41.0 & $-0.5$ & 3.2M \\
\hline
\multirow{5}{*}{OR1}
 & Uniform & 41.7 & -- & 1.4M \\
 & Algebra-biased & 42.1 & $+0.4$ & 1.4M \\
 & Geometry-biased & 41.3 & $-0.4$ & 1.5M \\
 & Number-Theory-biased & 41.4 & $-0.3$ & 1.4M \\
 & Combinatorics-biased & 41.7 & $0.0$ & 1.5M \\
\hline
\end{tabular}
}
\vspace{-2mm}
\end{table}

Tables~\ref{tab:validation_difficulty_bias} and~\ref{tab:validation_subdomain_bias} show relatively small accuracy changes under the tested bias conditions.
The reported two-benchmark average changes range from $-0.4$ to $+0.2$ percentage points under difficulty bias, while OlympiadBench changes range from $-0.5$ to $+0.4$ points under subdomain bias.
Rollout counts also vary with the validation composition.
These observations are limited to the tested settings, and the comparisons may additionally reflect test-set sampling differences across conditions.

\subsection{Code Generation with Full and Partial Verifiers}
\label{app:code_generation}

To evaluate L2RR beyond mathematical reasoning, we train Qwen2.5-Coder-1.5B-Instruct on the training split of CodeContests~\citep{li2022alphacode}, whose problems include executable test cases, and use HumanEval+ and MBPP+~\citep{liu2023evalplus} as target benchmarks.
We reserve 10\% of each target benchmark for sampler feedback and checkpoint selection and evaluate on the disjoint remaining 90\%.
Target validation examples are not included in L2RR's policy training, and AR3PO uses the same validation splits for checkpoint selection.

We compare two verifier conditions: full unit tests and a partial verifier that randomly selects 50\% of the available tests at each call.
In the partial-verifier condition, both training rewards and target feedback use the sampled test subsets.
Final evaluation always uses the full test suites for the held-out test problems, regardless of the verifier used during training.

Table~\ref{tab:code_generation_verifiers} shows that L2RR improves the average score over AR3PO by 1.3 percentage points with the full verifier and by 1.7 points with the partial verifier.
L2RR uses 2.7M and 2.6M rollouts in these settings, respectively, compared with 2.5M for AR3PO.
These results provide preliminary evidence that the L2RR training mechanism is applicable to code generation under the tested verifier conditions.

\begin{table}[t]
\centering
\caption{\textbf{Code generation with full and partial verifiers.} We report test scores (\%) on HumanEval+ and MBPP+, their average, and cumulative source-rollout counts for Qwen2.5-Coder-1.5B-Instruct trained on CodeContests. The verifier condition applies to training and target feedback, while final evaluation always uses the full test suites.}
\label{tab:code_generation_verifiers}
\setlength{\tabcolsep}{7pt}
\renewcommand{\arraystretch}{1.1}
\resizebox{0.9\textwidth}{!}{
\begin{tabular}{l|c|ccc|c}
\hline
Verifier & Method & HumanEval+ & MBPP+ & Avg. & Total rollouts \\
\hline\hline
\multirow{2}{*}{Full unit tests}
 & AR3PO & 63.3 & 58.5 & 60.9 & 2.5M \\
 & L2RR & \textbf{64.9} & \textbf{59.4} & \textbf{62.2} & 2.7M \\
\hline
\multirow{2}{*}{Random 50\% of tests}
 & AR3PO & 59.8 & 55.6 & 57.7 & 2.5M \\
 & L2RR & \textbf{61.9} & \textbf{56.9} & \textbf{59.4} & 2.6M \\
\hline
\end{tabular}
}
\vspace{-2mm}
\end{table}

\begin{table}[t]
\centering
\caption{\textbf{Performance (\%) comparison across six math reasoning benchmarks.}
Three models are trained on DM or OR1.
Base rows use target-validation checkpoint selection; +Target-Mix also trains on target validation examples; -TargetVal omits target-validation selection.
Total RO denotes cumulative source responses in millions; RO ratio compares counts with DAPO, not wall-clock time. All entries are single runs; five-seed means of the base methods and L2RR are in Table~\ref{tab:perf_rollouts}.
}
\label{tab:perf_rollouts_all}
\setlength{\tabcolsep}{5pt}
\renewcommand{\arraystretch}{1.12}
\resizebox{0.92\textwidth}{!}{
\begin{tabular}{c|l|cccccc|c|c|c}
\hline
Dataset & Method & Math500 & AIME24 & AMC & Gaokao & Miner. & Olymp. & Avg. & \# Rollout & RO ratio \\
\hline\hline

\multicolumn{11}{c}{\textit{Qwen2.5-Math-1.5B}}\\
\hline
\multirow{10}{*}{DM}
& DAPO  & {77.3} & {16.7} & {61.8} & {64.2} & {31.8} & {38.7} & {48.4} & 7.6M & 1.0$\times$ \\
& DAPO + Target-Mix & 77.9 & 17.6 & \textbf{62.2} & 65.1 & 32.6 & 39.4 & {49.1} & 8.2M & 0.9$\times$ \\
& DAPO - TargetVal  & 77.1 & 15.7 & 61.5 & 63.8 & 31.2 & 38.1 & {47.9} & 7.6M & 1.0$\times$ \\
& GRESO & {76.6} & {14.8} & {61.8} & {66.2} & {33.3} & {38.5} & {48.5} & 3.3M & 2.3$\times$ \\
& GRESO + Target-Mix & 77.4 & 15.7 & \textbf{62.2} & \textbf{67.3} & 34.5 & 39.4 & {49.4} & 3.8M & 2.0$\times$ \\
& GRESO - TargetVal  & 76.1 & {13.9} & 61.1 & 65.6 & 32.4 & 37.8 & {47.8} & 3.3M & 2.3$\times$ \\
& AR3PO & {77.8} & {17.6} & {61.1} & {65.4} & {35.8} & {39.0} & {49.5} & 2.5M & 3.0$\times$ \\
& AR3PO + Target-Mix & 78.9 & \textbf{18.5} & \textbf{62.2} & 66.5 & \textbf{36.2} & 39.9 & 50.4 & 2.8M & 2.7$\times$ \\
& AR3PO - TargetVal  & 76.9 & 16.7 & 60.8 & 64.8 & 35.1 & 38.2 & {48.8} & 2.5M & 3.0$\times$ \\
& L2RR  & {\textbf{79.9}} & {\textbf{18.5}} & {\textbf{62.2}} & {\textbf{67.3}} & {35.7} & {\textbf{41.5}} & {\textbf{50.9}} & 3.0M & 2.5$\times$ \\
\cline{1-11}
\multirow{10}{*}{OR1}
& DAPO  & {77.1} & {16.7} & {50.3} & {65.5} & {30.9} & {39.7} & {46.7} & 3.8M & 1.0$\times$ \\
& DAPO + Target-Mix & 77.8 & 17.6 & 51.4 & 66.4 & \textbf{31.6} & 40.6 & {47.6} & 4.0M & 1.0$\times$ \\
& DAPO - TargetVal  & 76.4 & 15.7 & {48.6} & 60.0 & 30.2 & 39.3 & {45.0} & 3.8M & 1.0$\times$ \\
& GRESO & {76.1} & {20.4} & {50.7} & {65.1} & {30.0} & {39.2} & {46.9} & 1.6M & 2.4$\times$ \\
& GRESO + Target-Mix & 76.8 & 20.4 & 51.4 & 65.9 & 30.8 & 39.8 & {47.5} & 1.8M & 2.1$\times$ \\
& GRESO - TargetVal  & 75.4 & {19.4} & 49.0 & 64.4 & 29.2 & 38.7 & {46.0} & 1.6M & 2.4$\times$ \\
& AR3PO & {76.5} & {20.4} & {51.0} & {65.0} & {30.2} & {39.6} & {47.1} & 1.4M & 2.7$\times$ \\
& AR3PO + Target-Mix & 77.1 & 20.4 & 51.4 & 65.9 & 30.9 & 40.3 & {47.7} & 1.7M & 2.2$\times$ \\
& AR3PO - TargetVal  & 76.1 & {19.4} & {50.3} & 64.3 & 29.7 & 38.9 & {46.5} & 1.4M & 2.7$\times$ \\
& L2RR  & {\textbf{78.8}} & {\textbf{21.3}} & {\textbf{52.4}} & {\textbf{66.7}} & {31.5} & {\textbf{41.7}} & {\textbf{48.7}} & 1.4M & 2.7$\times$ \\
\hline\hline

\multicolumn{11}{c}{\textit{DeepSeek-R1-Distill-Qwen-1.5B}}\\
\hline
\multirow{10}{*}{DM}
& DAPO  & {87.9} & {37.0} & {71.6} & {78.7} & {35.3} & {55.9} & {61.1} & 2.4M & 1.0$\times$ \\
& DAPO + Target-Mix & 88.5 & 37.0 & 72.3 & 79.1 & 35.6 & 56.4 & {61.5} & 2.8M & 0.9$\times$ \\
& DAPO - TargetVal  & 87.3 & 36.1 & 71.3 & 78.1 & 34.8 & 55.3 & 60.5 & 2.4M & 1.0$\times$ \\
& GRESO & {87.7} & {37.0} & {71.3} & {78.4} & {33.9} & {55.1} & {60.6} & 1.6M & 1.5$\times$ \\
& GRESO + Target-Mix & 88.6 & \textbf{38.0} & 72.0 & 78.9 & 34.5 & 56.0 & 61.3 & 1.8M & 1.3$\times$ \\
& GRESO - TargetVal  & 87.2 & 37.0 & 70.9 & 77.9 & 33.6 & 54.6 & {60.2} & 1.6M & 1.5$\times$ \\
& AR3PO & {88.0} & {37.0} & {71.3} & {78.6} & {34.8} & {55.6} & {60.9} & 1.7M & 1.4$\times$ \\
& AR3PO + Target-Mix & 88.9 & 37.0& 72.3 & 79.4 & 35.5 & 56.4 & {61.6} & 1.9M & 1.3$\times$ \\
& AR3PO - TargetVal  & 87.4 & 36.1 & 71.3 & 78.1 & 34.3 & 55.2 & {60.4} & 1.7M & 1.4$\times$ \\
& L2RR  & {\textbf{89.3}} & {\textbf{38.0}} & {\textbf{73.0}} & {\textbf{80.1}} & {\textbf{35.8}} & {\textbf{58.2}} & {\textbf{62.4}} & 1.6M & 1.5$\times$ \\
\cline{1-11}
\multirow{10}{*}{OR1}
& DAPO  & 84.8 & {25.0} & {68.2} & {74.0} & {34.1} & {54.2} & {56.7} & 2.4M & 1.0$\times$ \\
& DAPO + Target-Mix & 85.3 & 25.0 & \textbf{68.6} & 74.9 & 34.8 & 54.8 & {57.2} & 2.8M & 0.9$\times$ \\
& DAPO - TargetVal  & 84.2 & {24.1} & 67.9 & 73.4 & 33.5 & 53.6 & 56.1 & 2.4M & 1.0$\times$ \\
& GRESO & {85.9} & {26.9} & {66.9} & {75.2} & {33.6} & {55.5} & {57.3} & 1.5M & 1.6$\times$ \\
& GRESO + Target-Mix & 86.5 & 27.8 & 67.9 & 75.9 & 34.2 & 56.1 & {58.1} & 1.8M & 1.3$\times$ \\
& GRESO - TargetVal  & 85.4 & {25.9} & 66.2 & 74.8 & 33.0 & 55.1 & {56.7} & 1.5M & 1.6$\times$ \\
& AR3PO & {85.4} & {26.9} & {67.2} & {75.0} & {33.7} & {55.3} & {57.3} & 1.6M & 1.5$\times$ \\
& AR3PO + Target-Mix & 86.2 & 27.8 & 68.2 & 75.9 & 34.4 & 56.2 & {58.1} & 1.9M & 1.3$\times$ \\
& AR3PO - TargetVal  & 84.8 & {25.9} & 66.2 & 74.1 & 33.0 & 54.2 & 56.4 & 1.6M & 1.5$\times$ \\
& L2RR  & {\textbf{86.8}} & {\textbf{28.7}} & {68.2} & {\textbf{76.7}} & {\textbf{35.1}} & {\textbf{59.1}} & {\textbf{59.1}} & 1.5M & 1.6$\times$ \\
\hline\hline

\multicolumn{11}{c}{\textit{Qwen2.5-Math-7B}}\\
\hline
\multirow{10}{*}{DM}
& DAPO  & {82.9} & {34.3} & {79.1} & {71.7} & {35.4} & {43.6} & {57.8} & 13.1M & 1.0$\times$ \\
& DAPO + Target-Mix & 83.5 & \textbf{35.2} & 79.7 & 72.3 & 36.1 & 44.2 & 58.5 & 13.7M & 1.0$\times$ \\
& DAPO - TargetVal  & 82.1 & 33.3 & 78.4 & 71.0 & 34.8 & 43.0 & 57.1 & 13.1M & 1.0$\times$ \\
& GRESO & {82.2} & {32.4} & {80.7} & {70.2} & {35.4} & {44.1} & {57.5} & 6.3M & 2.1$\times$ \\
& GRESO + Target-Mix & 83.0 & 33.3 & \textbf{81.8} & 71.1 & 36.2 & 44.9 & 58.4 & 6.7M & 2.0$\times$ \\
& GRESO - TargetVal  & 81.6 & 31.5 & 80.1 & 69.3 & 34.8 & 43.4 & 56.8 & 6.3M & 2.1$\times$ \\
& AR3PO & {82.5} & {33.3} & {80.1} & {71.0} & {35.6} & {43.6} & {57.7} & 6.6M & 2.0$\times$ \\
& AR3PO + Target-Mix & 83.4 & \textbf{35.2} & 81.4 & 72.1 & 35.9 & 44.4 & {58.7} & 6.9M & 1.9$\times$ \\
& AR3PO - TargetVal  & 82.0 & 32.4 & 79.4 & 70.1 & 34.7 & 42.8 & {56.9} & 6.6M & 2.0$\times$ \\
& L2RR  & {\textbf{83.7}} & {34.3} & {\textbf{81.8}} & {\textbf{72.5}} & {\textbf{36.8}} & {\textbf{46.8}} & {\textbf{59.3}} & 6.0M & 2.2$\times$ \\
\cline{1-11}
\multirow{10}{*}{OR1}
& DAPO  & {82.9} & {34.3} & {63.2} & {67.3} & {34.9} & {46.3} & 54.8 & 11.4M & 1.0$\times$ \\
& DAPO + Target-Mix & 82.9 & 35.2 & 63.9 & 67.6 & 35.6 & 46.9 & {55.4} & 11.8M & 1.0$\times$ \\
& DAPO - TargetVal  & 82.1 & 33.3 & 62.2 & 66.5 & 33.6 & 45.8 & 53.9 & 11.4M & 1.0$\times$ \\
& GRESO & {82.3} & {35.2} & {64.5} & {66.8} & {36.5} & {45.7} & {55.2} & 3.4M & 3.4$\times$ \\
& GRESO + Target-Mix & 83.1 & \textbf{36.1} & 65.5 & 67.5 & 37.1 & 46.2 & 55.9 & 3.7M & 3.1$\times$ \\
& GRESO - TargetVal  & 83.4 & {34.3} & 63.5 & 65.8 & 35.8 & 45.0 & 54.6 & {3.4M} & {3.4$\times$} \\
& AR3PO & {82.4} & {\textbf{36.1}} & {64.5} & {66.5} & {36.6} & {45.7} & {55.3} & 3.6M & 3.2$\times$ \\
& AR3PO + Target-Mix & 83.2 & \textbf{36.1} & 65.5 & 67.3 & 37.2 & 46.5 & {56.0} & 3.8M & 3.0$\times$ \\
& AR3PO - TargetVal  & 81.6 & 35.2 & 63.5 & 65.7 & 35.9 & 45.0 & {54.5} & 3.6M & 3.2$\times$ \\
& L2RR  & {\textbf{83.6}} & {\textbf{36.1}} & {\textbf{65.9}} & {\textbf{67.9}} & {\textbf{37.6}} & {\textbf{49.7}} & \textbf{{56.8}} & 3.3M & 3.5$\times$ \\
\hline
\end{tabular}
}
\end{table}

\section{Related Work}
\label{sec:related}

\textbf{Reinforcement learning from verifiable rewards.}
RLVR has emerged as a powerful approach for improving the reasoning ability of LLMs, which leverages verifiers to provide rule based outcome rewards such as correctness in mathematics.
Methods such as \emph{proximal policy optimization}~{(PPO)}~\citep{ouyang2022training} and GRPO~\citep{shao2024deepseekmath} are representative algorithms.
Encouraged by their success, a growing body of work has emerged to further improve them for LLM reasoning~\citep{yuan2025s,yu2025dapo,liu2025understanding,xiong2025minimalist,zhang2025improving,zheng2025act}.
For example, the decoupled clip and dynamic sampling policy optimization algorithm (DAPO) proposed by~\cite{yu2025dapo} introduces several techniques to improve GRPO, including dynamic sampling, which filters out zero-variance prompts and refills the training batch with effective training data through resampling.
\cite{zhang2025improving} proposed the adaptive rollout and response reuse policy optimization algorithm (AR3PO) that improves RLVR sampling efficiency via adaptive rollouts and response reuse, with corrections to reduce mismatch between the current policy and the behavior policy that generated the reused data.
GRPO with efficient selective rollout method (GRESO)~\citep{zheng2025act} reduces rollout cost by predicting and skipping prompts that are likely to stay uninformative, leveraging the empirical temporal consistency of zero-advantage prompts across epochs. Guided adversarial self-play (GASP) improves reasoning robustness to corrupted context through guided adversarial self-play~\citep{li2026learning}. Consolidating rewarded perturbations (CoRP) explores gradient-free LLM post-training by consolidating rewarded parameter perturbations into a single model~\citep{zhang2026consolidating}.

A related line of work studies data selection to improve the efficiency of supervised fine-tuning for LLMs~\citep{ivison2025large,xia2024less,muennighoff2025s1}, and parallel efforts target data efficiency in RL for LLMs~\citep{muldrew2024active,li2025limr,fatemi2025concise}.
These approaches often require substantial training on the full dataset to identify useful samples, so they provide limited end-to-end savings in rollout-heavy RL training. More broadly, \cite{du2024AUF} make an important advance in non-stationary decision-making through the first polynomial-time rehearsal-based method for their linear-Gaussian AUF formulation. Their work shares our emphasis on cost-aware decision-making, while focuses on minimizing action costs to avoid undesired outcomes rather than allocating rollouts for RLVR training.

\textbf{Bi-level optimization for sample reweighting.}
A closely related line of research studied bi-level optimization for learning data weights: an inner loop trains model parameters on weighted training samples, while an outer loop updates the weights to minimize loss on a held-out development set.
Early hypergradient methods differentiate through the training dynamics to optimize such outer objectives~\citep{maclaurin2015gradient,franceschi2018bilevel}.
Building on this paradigm, \cite{ren2018learning} proposed to learn per-example weights so that one gradient step on the reweighted training loss improves a clean validation objective, and \cite{shu2019meta} learned an explicit loss-to-weight mapping using a small meta set.
Conceptually, these methods use limited target-side feedback to steer which training signals matter most; our setting is analogous in spirit, but differs technically because RLVR with GRPO induces history-dependent degeneracy, so we need to consider rollout efficiency in addition to validation performance.

On the theory side, \cite{nemirovski2009robust} established the foundational stochastic mirror-descent analyses.
\cite{duchi2011adaptive} provided regret and convergence guarantees for stochastic first-order methods, which motivates analyzing our stochastic sampler update via standard gradient-mapping machinery.
Recently, \cite{khanduri2021near} established convergence and complexity guarantees for stochastic bilevel optimization under smoothness and strong-convexity-type conditions, under their stated assumptions; these results do not by themselves prove the validity of the present one-step RLVR surrogate.


\section{Proof of Theorem~\ref{thm:informal_main}}
\label{app:proofs}

We first prove the sampler's online regret bound, then examine the effects of estimation error and rollout allocation. Finally, we give a separate target-loss guarantee under additional assumptions.

\subsection{Preliminary}

Let $\mathcal F_t$ contain the training history before round $t$, including $\q_t$, $\theta_t$, and the counters $z_{i,t}$. The feedback used in Eq.~\eqref{eq:q_update_impl} is
\[
\widehat g_t=\widehat g_t^{\rmtgt}+\lambda_T\psi_t,
\qquad \psi_{t,i}=1-p_e^{z_{i,t}}\in[0,1],
\]
where $\psi_t$ is $\mathcal F_t$-measurable and $\lambda_T$ is fixed within the run. The feedback includes batch sampling, rollout generation, and any target-side estimation randomness. Repeated prompt occurrences, if allowed, contribute to the same coordinate of $\widehat g_t^{\rmtgt}$. Starting from $q_{1,i}=1/N_S$, exponentiated gradient gives
\begin{equation}
\label{eq:q_update_thm}
q_{t+1,i}=\frac{q_{t,i}\exp(-\eta_q\widehat g_{t,i})}
{\sum_j q_{t,j}\exp(-\eta_q\widehat g_{t,j})}.
\end{equation}
For finite feedback and $\eta_q>0$, all coordinates remain strictly positive.

After the feedback is realized, define the linear losses
$\widehat\ell_t(u)=\langle\widehat g_t,u\rangle$ and
$\widehat\ell_t^{\rmtgt}(u)=\langle\widehat g_t^{\rmtgt},u\rangle$ on $\DeltaS$.
Both average regrets, defined as in Eq.~\eqref{eq:linear_regret}, compare against the best fixed distribution on the same realized sequence. This is external regret for feedback losses, not the difference between training trajectories induced by different samplers.

Define the conditional mean and centered error by
\[
m_t=\E[\widehat g_t\mid\mathcal F_t],\qquad
\xi_t=\widehat g_t-m_t,
\]
and set
\begin{equation}
\label{eq:sigma_q_def}
\sigma_{q,t}^2=\E\|\xi_t\|_2^2,
\qquad \bar\sigma_q^2=\frac1T\sum_{t=1}^T\sigma_{q,t}^2.
\end{equation}

\begin{myAssum}[Feedback second moment]
\label{ass:feedback_moment}
The feedback is square-integrable and satisfies
\[
\frac1T\sum_{t=1}^T\E\|\widehat g_t\|_\infty^2\le H^2,
\]
where $H$ is independent of the horizon. This is a condition on the feedback, not a consequence of normalized sampling weights: their denominators can approach zero without an enforced sampler floor.
\end{myAssum}

\subsection{Main Proof}

\begin{myLemma}[One-step entropy inequality]
\label{lem:eg_regret_step}
For any $u\in\DeltaS$, finite feedback, and $\eta_q>0$,
\[
\langle\widehat g_t,\q_t-u\rangle
\le\frac{\KL(u\|\q_t)-\KL(u\|\q_{t+1})}{\eta_q}
+\frac{\eta_q}{2}\|\widehat g_t\|_\infty^2.
\]
\end{myLemma}
\begin{proof}
Let $h(q)=\sum_iq_i\log q_i$ and $p=\q_{t+1}$. The update is the minimizer of
$\langle\widehat g_t,p\rangle+\eta_q^{-1}\KL(p\|\q_t)$.
Its optimality condition and the Bregman three-point identity give
\[
\eta_q\langle\widehat g_t,p-u\rangle
\le \KL(u\|\q_t)-\KL(u\|p)-\KL(p\|\q_t).
\]
Add $\eta_q\langle\widehat g_t,\q_t-p\rangle$ and use Pinsker's inequality and H\"older's inequality. Writing $d=\|\q_t-p\|_1$, the remaining terms satisfy
\[
\eta_q\langle\widehat g_t,\q_t-p\rangle-\KL(p\|\q_t)
\le\eta_q\|\widehat g_t\|_\infty d-\tfrac12d^2
\le\tfrac12\eta_q^2\|\widehat g_t\|_\infty^2.
\]
Division by $\eta_q$ proves the claim. Boundary comparators are allowed with the convention $0\log0=0$.
\end{proof}

\begin{myThm}[Formal statement of Theorem~\ref{thm:informal_main}]
\label{thm:main}
For any realized finite feedback sequence, any constant $\eta_q>0$, and uniform $\q_1$,
\begin{equation}
\label{eq:regret_pathwise}
\widehat R_T\le\frac{\log N_S}{\eta_q T}
+\frac{\eta_q}{2T}\sum_{t=1}^T\|\widehat g_t\|_\infty^2,
\qquad
\widehat R_T^{\rmtgt}\le\widehat R_T+\lambda_T.
\end{equation}
Under Assumption~\ref{ass:feedback_moment}, setting $\eta_q=a/\sqrt T$ gives Eqs.~\eqref{eq:sampler_regret_main}--\eqref{eq:target_feedback_regret_main}. If $\lambda_T\le c_\lambda/\sqrt T$, then
\[
\E\widehat R_T^{\rmtgt}
\le \frac{\log N_S/a+aH^2/2+c_\lambda}{\sqrt T}.
\]
\end{myThm}
\begin{proof}
Sum Lemma~\ref{lem:eg_regret_step} and telescope the KL terms. For every $u\in\DeltaS$, uniform initialization gives
$\KL(u\|\q_1)=\log N_S+\sum_i u_i\log u_i\le\log N_S$,
and the final KL divergence is nonnegative. The bound holds simultaneously for all $u$, so it also holds for the best comparator chosen after the full feedback sequence is observed. Dividing by $T$ proves the first inequality in Eq.~\eqref{eq:regret_pathwise}.

For each comparator $u$,
\begin{align*}
\frac1T\sum_{t=1}^T\langle\widehat g_t^{\rmtgt},\q_t-u\rangle
&=\frac1T\sum_{t=1}^T\langle\widehat g_t,\q_t-u\rangle
+\frac{\lambda_T}{T}\sum_{t=1}^T\langle\psi_t,u-\q_t\rangle\\
&\le\widehat R_T+\lambda_T,
\end{align*}
since both $\langle\psi_t,u\rangle$ and $\langle\psi_t,\q_t\rangle$ lie in $[0,1]$. Maximizing over $u$ proves the second inequality. Taking expectations, applying the second-moment assumption, and substituting the step size proves the rates. The argument is pathwise before expectations and allows feedback that depends on the sampler and training history; unbiasedness and PL conditions are not needed.
\end{proof}

\paragraph{How noise enters the regret bound.}
The martingale-difference property yields the exact decomposition
\begin{equation}
\label{eq:regret_second_moment}
\E\|\widehat g_t\|_2^2=\E\|m_t\|_2^2+\sigma_{q,t}^2.
\end{equation}
Since $\|\widehat g_t\|_\infty\le\|\widehat g_t\|_2$, one may replace $H^2$ in the expected regret bound by
$T^{-1}\sum_t\E\|m_t\|_2^2+\bar\sigma_q^2$.
Reducing the variance tightens this upper bound when the other terms are held fixed. The regret is for the realized feedback losses; it does not require a vanishing bias or policy-tracking error.

\subsection{Assumptions}

For the separate target-loss analysis, we define the outer objective for a fixed sampler $\q\in\DeltaS$:
\begin{equation}
\label{eq:Phi_def_thm}
\Phi(\q)
=
\E[L_{\rmtgt}\!\big(\theta_{T+1}(\q)\big)]
+
\lambda\,L_{\adv}(\q),
\end{equation}
where $\theta_{T+1}(\q)$ is a random parameter after $T$ GRPO updates when sampling data from $\q$, and $L_{\adv}(\q)$ is the expected history-based time-averaged rollout-waste functional. This defines a deterministic idealized fixed-sampler objective. Relating the coupled algorithm to it requires the assumptions below; the relation is not implied by the one-step surrogate.

Define the KL mirror step
\[
\mathcal{M}_{\eta_q}(\q;v)
=
\arg\min_{u\in\DeltaS}
\left\{
\langle v,u\rangle+\frac{1}{\eta_q}\KL(u\|\q)
\right\},
\]
and the corresponding KL-gradient mapping of $\Phi$:
\begin{equation}
\label{eq:grad_mapping_thm}
G_{\eta_q}(\q)
=
\frac{1}{\eta_q}\Big(\q-\mathcal{M}_{\eta_q}(\q;\nabla\Phi(\q))\Big).
\end{equation}

\begin{myAssum}
\label{ass:smooth}
Assume $\Phi$ is continuously differentiable on a neighborhood of $\DeltaS$ and its gradient is $L_q$-Lipschitz on the simplex:
\[
\|\nabla\Phi(\q)-\nabla\Phi(\q')\|_2
\le
L_q\|\q-\q'\|_2,
\qquad
\forall \q,\q'\in\DeltaS.
\]
Here $L_q>0$ is an assumed smoothness constant, not a consequence of the normalized one-step weights. No fixed positive sampler floor is enforced by Eq.~\eqref{eq:q_update_thm}. Let $\Phi^{\inf}=\inf_{q\in\DeltaS}\Phi(q)>-\infty$. Define $M_\Phi=\sup_{\q\in\DeltaS}\|\nabla\Phi(\q)\|_2$ and $K_\Phi=L_q+3M_\Phi$. Compactness and the global Lipschitz-gradient condition imply $M_\Phi<\infty$.
\end{myAssum}

\begin{myAssum}
\label{ass:unbiased}
Let $\widehat g_t=(\widehat g_{t,i})_{i=1}^{N_S}$ be the stochastic gradient used in~\eqref{eq:q_update_thm}.
In addition to the preceding online setup, assume
\[
\E[\widehat g_t\mid\mathcal F_t]=\nabla\Phi(\q_t).
\]
This identifies $m_t$ with $\nabla\Phi(\q_t)$ for the target-loss analysis only; it is not required for Theorem~\ref{thm:main}.
\end{myAssum}

\begin{myAssum}
\label{ass:bias}
Define the meta objective
\[
F_\lambda(q) = L_{\mathrm{tgt}}(\theta^\star(q)) + \lambda L_{\mathrm{ws}}(q),
\]
where $\theta^\star(q)$ is a fixed, deterministic choice of idealized inner solution for sampler $q$. Write $f(q)=L_{\rmtgt}(\theta^\star(q))$ and $R(q)=L_{\mathrm{ws}}(q)\in[0,1]$, using the expected waste functional under constant sampling. This $R(q)$ is distinct from the instantaneous history-dependent linear penalty $\langle\psi_t,q\rangle$. Assume $F_\lambda$ is differentiable and the mismatch between the two reference-objective gradients is bounded in second moment:
\[
\mathbb{E}\left[
\left\|
\nabla \Phi(q_t) - \nabla F_\lambda(q_t)
\right\|_2^2
\right]
\le \varepsilon^2,
\qquad \forall t \in [T].
\]
\end{myAssum}

\begin{myAssum}
\label{ass:tracking}
Assume along the coupled trajectory $(\theta_t,\q_t)$,
\[
\E\!\left[\|\theta_t-\theta^\star(\q_t)\|_2^2\right]\le \delta_t,
\]
and $L_{\rmtgt}$ is Lipschitz in $\theta$ on the trajectory: for some $L_\theta>0$,
\[
\left|L_{\rmtgt}(\theta)-L_{\rmtgt}(\theta^\star(\q))\right|
\le
L_\theta\|\theta-\theta^\star(\q)\|_2.
\]
\end{myAssum}

\begin{myAssum}
\label{ass:PL}
Assume the minimum of $F_\lambda$ on $\DeltaS$ is attained, and let $\q_\lambda^\star\in\arg\min_{\q\in\DeltaS}F_\lambda(\q)$.
Assume that on the region visited by $\{\q_t\}$,
$F_\lambda$ satisfies a PL inequality with respect to its KL-gradient mapping
$G^{F_\lambda}_{\eta_q}$:
there exists $\mu>0$ such that
\[
F_\lambda(\q)-F_\lambda(\q_\lambda^\star)
\le
\frac{1}{2\mu}\left\|G^{F_\lambda}_{\eta_q}(\q)\right\|_2^2.
\]
\end{myAssum}

The mapping $G^{F_\lambda}_{\eta_q}$ is defined by replacing $\Phi$ with $F_\lambda$ in Eq.~\eqref{eq:grad_mapping_thm}.

\subsection{Technical Lemmas}

To distinguish feedback regret from an idealized comparison, let $v_t$ be an $\mathcal F_t$-measurable reference target-gradient vector and write
\[
\E[\widehat g_t^{\rmtgt}\mid\mathcal F_t]=v_t+b_t,
\qquad \E\|b_t\|_\infty\le\epsilon_t.
\]
Define the cumulative reference regret $R_T^{\mathrm{ref}}=\max_{u\in\DeltaS}\sum_t\langle v_t,\q_t-u\rangle$. The reference vectors are evaluated along the actual history; choosing $v_t=\nabla f(\q_t)$ does not make this regret in the values of a nonconvex $f$.

\begin{myProp}[Reference-gradient regret]
\label{prop:reference_regret}
Under Assumption~\ref{ass:feedback_moment} and the preceding bias bound,
\begin{equation}
\label{eq:reference_regret}
\frac{\E R_T^{\mathrm{ref}}}{T}
\le\frac{\log N_S/a+aH^2/2+\sqrt{\bar\sigma_q^2}}{\sqrt T}
+\lambda_T+\frac2T\sum_{t=1}^T\epsilon_t.
\end{equation}
\end{myProp}
\begin{proof}
Because the waste gradient is $\mathcal F_t$-measurable,
$\widehat g_t=v_t+b_t+\lambda_T\psi_t+\xi_t$.
For every $u$, decompose the reference regret into these four terms. Use $\|\q_t-u\|_1\le2$ for the bias and $\langle\psi_t,u-\q_t\rangle\le1$ for the penalty. Taking the supremum gives
\[
R_T^{\mathrm{ref}}
\le T\widehat R_T+\lambda_T T+2\sum_t\|b_t\|_\infty
-\sum_t\langle\xi_t,\q_t\rangle
+\left\|\sum_t\xi_t\right\|_\infty.
\]
Adaptedness implies $\E\langle\xi_t,\q_t\rangle=0$. Orthogonality of square-integrable martingale differences and Jensen's inequality give
\[
\E\left\|\sum_t\xi_t\right\|_\infty
\le\sqrt{\E\left\|\sum_t\xi_t\right\|_2^2}
=\sqrt{\sum_t\sigma_{q,t}^2}.
\]
Divide by $T$ and apply the average-regret bound in Theorem~\ref{thm:main}. The last term accounts for the comparator being selected in hindsight, so no independence of that comparator is assumed.
\end{proof}

With uniformly bounded $\bar\sigma_q^2$, $\lambda_T\le c_\lambda/\sqrt T$, and $\sum_t\epsilon_t=\O(\sqrt T)$, reference-gradient regret is also $\O(\sqrt T)$. A constant bias bound $\epsilon_t\le\epsilon$ instead contributes $2\epsilon T$ to cumulative regret, or $2\epsilon$ to average regret. Thus a constant residual in average regret is not a constant additive term in cumulative regret. For a fixed penalty weight, removing the penalty similarly costs at most $\lambda_T T$; no such removal is needed for the regularized feedback guarantee.

\begin{myLemma}[Noise-sensitive EG descent]
\label{lem:md_rate}
Under Assumptions~\ref{ass:smooth} and~\ref{ass:unbiased}, with finite $\sigma_{q,t}^2$ as in Eq.~\eqref{eq:sigma_q_def}, the EG iterates satisfy, for $0<\eta_q\le1/L_q$,
\begin{equation}
\label{eq:md_rate}
\min_{t\in[T]}\E\|G_{\eta_q}(\q_t)\|_2^2
\le\frac{2\Delta_\Phi}{\eta_q T}+K_\Phi\eta_q\bar\sigma_q^2,
\qquad \Delta_\Phi=\Phi(\q_1)-\Phi^{\inf}.
\end{equation}
\end{myLemma}
\begin{proof}
We exploit the softmax form of EG before bounding the noise. This preserves cancellation of the first-order, conditionally mean-zero error. Nonconvex stochastic mirror-descent rates accommodating entropy geometry are also studied in~\citep{fatkhullin2024taming}; the following proof is self-contained for this EG update.

\textbf{Step 1: a smooth function in logit coordinates.}
Let $S(z)=\operatorname{softmax}(z)$, $p=S(z)$, and $J(z)=\operatorname{diag}(p)-pp^\top$. Since $0\preceq J(z)\preceq I$, $S$ is 1-Lipschitz in $\ell_2$. For $p'=S(z')$,
\[
\|J(z)-J(z')\|_{\mathrm{op}}
\le\|p-p'\|_\infty+(\|p\|_2+\|p'\|_2)\|p-p'\|_2
\le3\|z-z'\|_2.
\]
Consequently $\Psi(z)=\Phi(S(z))$ has a $K_\Phi$-Lipschitz gradient: using $\nabla\Psi(z)=J(z)\nabla\Phi(S(z))$,
\[
\|\nabla\Psi(z)-\nabla\Psi(z')\|_2
\le (L_q+3M_\Phi)\|z-z'\|_2.
\]
This bound is uniform over finite logits and does not require a fixed $q_{\min}>0$.

\textbf{Step 2: use conditional unbiasedness before Young-type bounds.}
Set $a_t=\nabla\Phi(\q_t)$, $e_t=\widehat g_t-a_t$,
$z_t=\log\q_t-\eta_q a_t$, and $p_t=S(z_t)=\mathcal M_{\eta_q}(\q_t;a_t)$.
The actual iterate is $\q_{t+1}=S(z_t-\eta_q e_t)$. Conditional on $\mathcal F_t$, $z_t$ is fixed and $\E[e_t\mid\mathcal F_t]=0$. Smoothness of $\Psi$ therefore gives
\begin{align*}
\E[\Phi(\q_{t+1})\mid\mathcal F_t]
&\le\Phi(p_t)-\eta_q\langle\nabla\Psi(z_t),\E[e_t\mid\mathcal F_t]\rangle
 +\frac{K_\Phi\eta_q^2}{2}\E[\|e_t\|_2^2\mid\mathcal F_t]\\
&=\Phi(p_t)+\frac{K_\Phi\eta_q^2}{2}\E[\|e_t\|_2^2\mid\mathcal F_t].
\end{align*}

\textbf{Step 3: exact-step descent and telescoping.}
Negative entropy $h(p)=\sum_i p_i\log p_i$ is 1-strongly convex in $\ell_2$. The exact mirror-step optimality condition gives
\[
\langle a_t,p_t-\q_t\rangle
\le-\eta_q^{-1}\langle\nabla h(p_t)-\nabla h(\q_t),p_t-\q_t\rangle
\le-\eta_q^{-1}\|p_t-\q_t\|_2^2.
\]
By $L_q$-smoothness and $\eta_q\le1/L_q$,
\[
\Phi(p_t)\le\Phi(\q_t)-\frac{1}{2\eta_q}\|p_t-\q_t\|_2^2
=\Phi(\q_t)-\frac{\eta_q}{2}\|G_{\eta_q}(\q_t)\|_2^2.
\]
Combining the last two bounds, taking expectations, and summing yields
\[
\frac{\eta_q}{2}\sum_{t=1}^T\E\|G_{\eta_q}(\q_t)\|_2^2
\le\Delta_\Phi+\frac{K_\Phi\eta_q^2}{2}\sum_{t=1}^T\sigma_{q,t}^2.
\]
Divide by $\eta_qT/2$ and take the minimum. The first-order noise cancellation uses conditional unbiasedness; the bound does not apply unchanged to an arbitrary biased feedback vector.
\end{proof}

\begin{myCor}[Target-loss bound]
\label{cor:target_loss}
Under Assumptions~\ref{ass:smooth}, \ref{ass:unbiased}, \ref{ass:bias}, \ref{ass:tracking}, and~\ref{ass:PL}, with finite variances in Eq.~\eqref{eq:sigma_q_def}, fix a horizon $T\ge1$, choose
$\eta_q=1/(L_q\sqrt T)$ and $0\le\lambda=\lambda_T\le c_\lambda/\sqrt T$, and hold both values fixed throughout the run. Here $c_\lambda\ge0$ is independent of $T$.
Let $\bar t\in\arg\min_{t\in[T]}\E\|G_{\eta_q}(\q_t)\|_2^2$, $R=L_{\mathrm{ws}}\in[0,1]$, and
$\q_{\lambda_T}^\star\in\arg\min_\q F_{\lambda_T}(\q)$. Then
\begin{equation}
\label{eq:original_rate_recovered}
\begin{aligned}
\E\!\left[L_{\rmtgt}(\theta_{\bar t})-L_{\rmtgt}(\theta^\star(\q_{\lambda_T}^\star))\right]
\le{}&\frac{2L_q\Delta_\Phi/\mu+K_\Phi\bar\sigma_q^2/(\mu L_q)+c_\lambda}{\sqrt T}\\
&{}+\frac{\varepsilon^2}{\mu}+L_\theta\sqrt{\delta_{\bar t}}.
\end{aligned}
\end{equation}
\end{myCor}
\begin{proof}
For mirror steps from the same $\q$, strong convexity and the two optimality conditions imply
\[
\|\mathcal M_{\eta_q}(\q;v)-\mathcal M_{\eta_q}(\q;v')\|_2
\le\eta_q\|v-v'\|_2.
\]
Indeed, writing the two steps as $p,p'$, addition of their variational inequalities yields
$\|p-p'\|_2^2\le\eta_q\langle v'-v,p-p'\rangle$.
Assumption~\ref{ass:bias} therefore gives
\[
\E\|G^{F_\lambda}_{\eta_q}(\q_{\bar t})\|_2^2
\le2\E\|G_{\eta_q}(\q_{\bar t})\|_2^2+2\varepsilon^2.
\]
Combine Lemma~\ref{lem:md_rate} with Assumption~\ref{ass:PL} to obtain
\[
\E[F_\lambda(\q_{\bar t})-F_\lambda(\q_\lambda^\star)]
\le\frac{2\Delta_\Phi}{\mu\eta_qT}
+\frac{K_\Phi\eta_q\bar\sigma_q^2+\varepsilon^2}{\mu}.
\]
For $f(\q)=L_{\rmtgt}(\theta^\star(\q))$, use the exact identity
\[
f(\q)-f(\q_\lambda^\star)=F_\lambda(\q)-F_\lambda(\q_\lambda^\star)
+\lambda[R(\q_\lambda^\star)-R(\q)].
\]
Taking expectations first gives, for any $0<\eta_q\le1/L_q$,
\begin{equation}
\label{eq:target_bound_sharp}
\begin{aligned}
\E[f(\q_{\bar t})-f(\q_\lambda^\star)]
\le{}&\frac{2\Delta_\Phi}{\mu\eta_qT}
+\frac{K_\Phi\eta_q\bar\sigma_q^2+\varepsilon^2}{\mu}\\
&{}+\lambda\bigl(R(\q_\lambda^\star)-\E R(\q_{\bar t})\bigr).
\end{aligned}
\end{equation}
Because $0\le R\le1$ and $0\le\lambda=\lambda_T\le c_\lambda/\sqrt T$,
$\lambda_T(R(\q_{\lambda_T}^\star)-\E R(\q_{\bar t}))\le c_\lambda/\sqrt T$.
Finally, Assumption~\ref{ass:tracking} and Cauchy--Schwarz give
\[
\E[L_{\rmtgt}(\theta_{\bar t})-f(\q_{\bar t})]
\le L_\theta\E\|\theta_{\bar t}-\theta^\star(\q_{\bar t})\|_2
\le L_\theta\sqrt{\delta_{\bar t}}.
\]
Substitute $\eta_q=1/(L_q\sqrt T)$ and add this bound to obtain Eq.~\eqref{eq:original_rate_recovered}.
\end{proof}

\paragraph{Rate across training horizons.}
For an asymptotic family of runs, assume the smoothness, gradient, PL, and initial-gap constants are controlled uniformly: for example $0<L_-\le L_q\le L_+$, $\mu\ge\mu_0>0$, $M_\Phi\le M_0$, $\Delta_\Phi\le\Delta_0$, and $\bar\sigma_q^2\le\sigma_0^2$, with $L_\theta$ uniformly bounded. The PL inequality must hold for the step sizes used in this family. Equation~\eqref{eq:original_rate_recovered} then gives
$\O(T^{-1/2})+\O(\bar\sigma_q^2/\sqrt T)+\O(\varepsilon^2)+L_\theta\sqrt{\delta_{\bar t}}$.
Thus the regularization transfer is absorbed into the convergence term, with no new constant noise or regularization floor. The comparator is the minimizer of $F_{\lambda_T}$ for the same horizon.

\begin{myRemark}
Corollary~\ref{cor:target_loss} is a conditional optimization statement about an analytically selected iterate, separate from the online regret in Theorem~\ref{thm:main}. It does not identify the final or validation-selected checkpoint with that iterate. The fixed-sampler reference $\Phi$, conditional unbiasedness, gradient mismatch, tracking, and the mapping-based PL condition are additional assumptions, not consequences of the one-step implementation or of the feedback regret theorem. The weight $\lambda_T$ depends on the planned training horizon but is constant within each run. This theoretical parameter rule is not asserted for the reported experiments.
\end{myRemark}

\subsection{Rollout Allocation Proposition}

The following statements connect the rollout model to the variance in Eqs.~\eqref{eq:regret_second_moment} and~\eqref{eq:original_rate_recovered}. They are not needed for the feedback regret theorem.

\begin{myAssum}
\label{ass:rollout_variance}
At iteration $t$, the rollouts per prompt are $\{G_{i,t}\}_{i\in\B_t}$ and the batch-local budget is $\mathrm{RO}_t$, with $\sum_{i\in\B_t}G_{i,t}\le\mathrm{RO}_t$ and $0<G_{\min}\le G_{i,t}\le G$.
Condition on $\mathcal F_t$ and the sampled batch $\B_t$. For the coordinate-variance and allocation statements below, assume distinct sampled indices. For repeated indices, the same argument requires bounds for the aggregated coordinate contributions. The allocation and variance proxies are measurable given this history and batch, before the evaluated rollouts are generated. Assume the target-driven coordinate estimator satisfies
\begin{equation}
\label{eq:var_model_assump}
\mathrm{Var}\!\left(\widehat g^{\rmtgt}_{t,i}\mid\mathcal F_t,\B_t\right)
\le
\frac{c_{i,t}}{G_{i,t}},
\qquad
c_{i,t}\ge 0,
\end{equation}
and assume $\lambda_T\psi_{t,i}$ is $\mathcal F_t$-measurable, thus it contributes no conditional variance given this history. The $c_{i,t}$ must bound the aggregated coordinate errors; setting $c=w^2$ heuristically does not establish this assumption.
\end{myAssum}

\begin{myLemma}
\label{lem:variance_bound}
Under Assumption~\ref{ass:rollout_variance}, define $m_{t,\B}=\E[\widehat g_t\mid\mathcal F_t,\B_t]$ and
\[
V_t=\E\!\left[\|m_{t,\B}-m_t\|_2^2\right].
\]
Then the total noise satisfies
\[
\sigma_{q,t}^2\le\E\!\left[\sum_{i\in\operatorname{supp}(\B_t)}\frac{c_{i,t}}{G_{i,t}}\right]+V_t.
\]
\end{myLemma}
\begin{proof}
The law of total conditional variance gives
\[
\sigma_{q,t}^2=\E\|\widehat g_t-m_{t,\B}\|_2^2+
\E\|m_{t,\B}-m_t\|_2^2.
\]
Conditional on the batch and history, unsampled target-feedback coordinates are identically zero and the waste term is deterministic. The first term is therefore the sum of the conditional variances of the sampled coordinates, bounded by Assumption~\ref{ass:rollout_variance}. The second term is $V_t$ and includes batch-selection randomness over all coordinates. It cannot generally be omitted.
\end{proof}

The next lemma gives the optimum of the strictly positive, unconstrained continuous weighted proxy. Clipping or discretization is not covered by this lemma.

\begin{myLemma}
\label{lem:alloc}
Fix iteration $t$ and a realized batch $\B_t$.
Let weights $w_{i,t}>0$ and $c_{i,t}>0$ be given for $i\in\B_t$, with $\mathrm{RO}_t>0$.
Consider the continuous relaxation (ignoring clipping and integer constraints)
\[
\min_{\{G_{i,t}>0\}}
\ \sum_{i\in\B_t}\frac{w_{i,t}c_{i,t}}{G_{i,t}}
\quad\text{s.t.}\quad
\sum_{i\in\B_t}G_{i,t}=\mathrm{RO}_t.
\]
Then the unique optimizer is
\[
G_{i,t}^\star
=
\mathrm{RO}_t\cdot
\frac{\sqrt{w_{i,t}c_{i,t}}}{\sum_{j\in\B_t}\sqrt{w_{j,t}c_{j,t}}},
\qquad i\in\B_t,
\]
and the optimal value equals
\[
\sum_{i\in\B_t}\frac{w_{i,t}c_{i,t}}{G_{i,t}^\star}
=
\frac{\left(\sum_{i\in\B_t}\sqrt{w_{i,t}c_{i,t}}\right)^2}{\mathrm{RO}_t}.
\]
\end{myLemma}

\begin{proof}
Form the Lagrangian
\[
\mathcal{L}(\{G_{i,t}\},\nu)
=
\sum_{i\in\B_t}\frac{w_{i,t}c_{i,t}}{G_{i,t}}+\nu\left(\sum_{i\in\B_t}G_{i,t}-\mathrm{RO}_t\right).
\]
Stationarity for each $i\in\B_t$ gives
\[
\frac{\partial \mathcal{L}}{\partial G_{i,t}}
=
-\frac{w_{i,t}c_{i,t}}{G_{i,t}^2}+\nu
=
0
\quad\Longrightarrow\quad
G_{i,t}
=
\sqrt{\frac{w_{i,t}c_{i,t}}{\nu}}.
\]
Impose the budget constraint:
\[
\sum_{i\in\B_t}\sqrt{\frac{w_{i,t}c_{i,t}}{\nu}}
=
\mathrm{RO}_t
\quad\Longrightarrow\quad
\sqrt{\nu}
=
\frac{\sum_{i\in\B_t}\sqrt{w_{i,t}c_{i,t}}}{\mathrm{RO}_t}.
\]
Substituting back yields the stated optimizer. Strict convexity of $\sum_iw_{i,t}c_{i,t}/G_{i,t}$ for positive products establishes uniqueness.
Finally,
\[
\sum_{i\in\B_t}\frac{w_{i,t}c_{i,t}}{G_{i,t}^\star}
=
\sum_{i\in\B_t}\frac{w_{i,t}c_{i,t}}{\mathrm{RO}_t}\cdot
\frac{\sum_{j\in\B_t}\sqrt{w_{j,t}c_{j,t}}}{\sqrt{w_{i,t}c_{i,t}}}
=
\frac{\left(\sum_{i\in\B_t}\sqrt{w_{i,t}c_{i,t}}\right)^2}{\mathrm{RO}_t}.
\]
\end{proof}

\begin{myProp}[Continuous allocation and the unweighted variance term]
\label{prop:variance_scaling}
Under Lemmas~\ref{lem:variance_bound} and~\ref{lem:alloc}, if the continuous allocation $G_{i,t}^\star$ is used, with strictly positive weights and proxies and without clipping, then
\[
\sigma_{q,t}^2\le V_t+
\E\!\left[\frac{\sum_j\sqrt{w_{j,t}c_{j,t}}}{\mathrm{RO}_t}
\sum_i\sqrt{\frac{c_{i,t}}{w_{i,t}}}\right].
\]
This is an identity-based substitution for the unweighted variance term, not the smaller minimum of the weighted proxy. It does not prove the implemented clipped allocation minimizes total noise.
\end{myProp}
\begin{proof}
Substitute $G_{i,t}^\star=\mathrm{RO}_t\sqrt{w_{i,t}c_{i,t}}/\sum_j\sqrt{w_{j,t}c_{j,t}}$ into $\sum_i c_{i,t}/G_{i,t}^\star$ and apply Lemma~\ref{lem:variance_bound}.
\end{proof}

\end{document}